\documentclass[lettersize,journal]{IEEEtran}
\usepackage{amsmath,amsfonts}
\usepackage{algorithmic}
\usepackage{algorithm}
\usepackage{array}
\usepackage[caption=false,font=normalsize,labelfont=sf,textfont=sf]{subfig}
\usepackage{textcomp}
\usepackage{stfloats}
\usepackage{url}
\usepackage{verbatim}
\usepackage{graphicx}
\usepackage{cite}
\usepackage{orcidlink}
\hypersetup{hidelinks}

\newcommand{\etal}{\textit{et al.}}
\usepackage{multirow}
\usepackage{xcolor}

\begin{document}
\title{Regression of Dense Distortion Field from a Single Fingerprint Image}

\author{
	Xiongjun Guan$^{\orcidlink{0000-0001-8887-3735}}$, 
	Yongjie Duan$^{\orcidlink{0000-0003-3741-9596}}$,
	Jianjiang Feng$^{\orcidlink{0000-0003-4971-6707}}$, ~\IEEEmembership{Member, IEEE}, 
	and Jie Zhou$^{\orcidlink{0000-0001-7701-234X}}$, ~\IEEEmembership{Senior Member, IEEE}
	\IEEEcompsocitemizethanks{\IEEEcompsocthanksitem
		
		This work was supported in part by the National Natural Science Foundation of China under Grant 61976121.
		The authors are with Department of Automation, Tsinghua University, Beijing 100084, China (e-mail: gxj21@mails.tsinghua.edu.cn; dyj17@mails.tsinghua.edu.cn; jfeng@tsinghua.edu.cn; jzhou@tsinghua.edu.cn).
		
	}
}

\markboth{IEEE TRANSACTIONS ON INFORMATION FORENSICS AND SECURITY}%
{Guan \MakeLowercase{\textit{et al.}}: Regression of Dense Distortion Field from a Single Fingerprint Image}


\maketitle

\begin{abstract}
Skin distortion is a long standing challenge in fingerprint matching, which causes false non-matches. 
Previous studies have shown that the recognition rate can be improved by estimating the distortion field from a distorted fingerprint and then rectifying it into a normal fingerprint.
However, existing rectification methods are based on principal component representation of distortion fields, which is not accurate and are very sensitive to finger pose.
In this paper, we propose a rectification method where a self-reference based network is utilized to directly estimate the dense distortion field of distorted fingerprint instead of its low dimensional representation. 
This method can output accurate distortion fields of distorted fingerprints with various finger poses and distortion patterns.
We conducted experiments on FVC2004 DB1\_A, expanded Tsinghua Distorted Fingerprint database (with additional distorted fingerprints in diverse finger poses and distortion patterns) and a latent fingerprint database. 
Experimental results demonstrate that our proposed method achieves the state-of-the-art rectification performance in terms of distortion field estimation and rectified fingerprint matching.
\end{abstract}

\begin{IEEEkeywords}
Fingerprint, distortion rectification, deep learning.
\end{IEEEkeywords}

\section{Introduction}

Fingerprint is one of the most important and widely used biometric traits due to its easy collection process, persistence and uniqueness \cite{maltoni2022handbook}. 
With the development of sensor technology and recognition algorithms in the past few decades, fingerprint recognition technologies developed rapidly and have been applied in various fields such as criminal investigation, mobile payment, and access control. 
Most existing fingerprint recognition algorithms extract features based on ridge and minutiae information, which are then utilized for fingerprint matching \cite{cappelli2010minutia}.
Although these algorithms achieve satisfactory performance on high-quality fingerprint images, they often fail to identify severely distorted fingerprints \cite{cappelli2006performance}. 

\begin{figure}[!t]
\centering
\includegraphics[width=.95\linewidth]{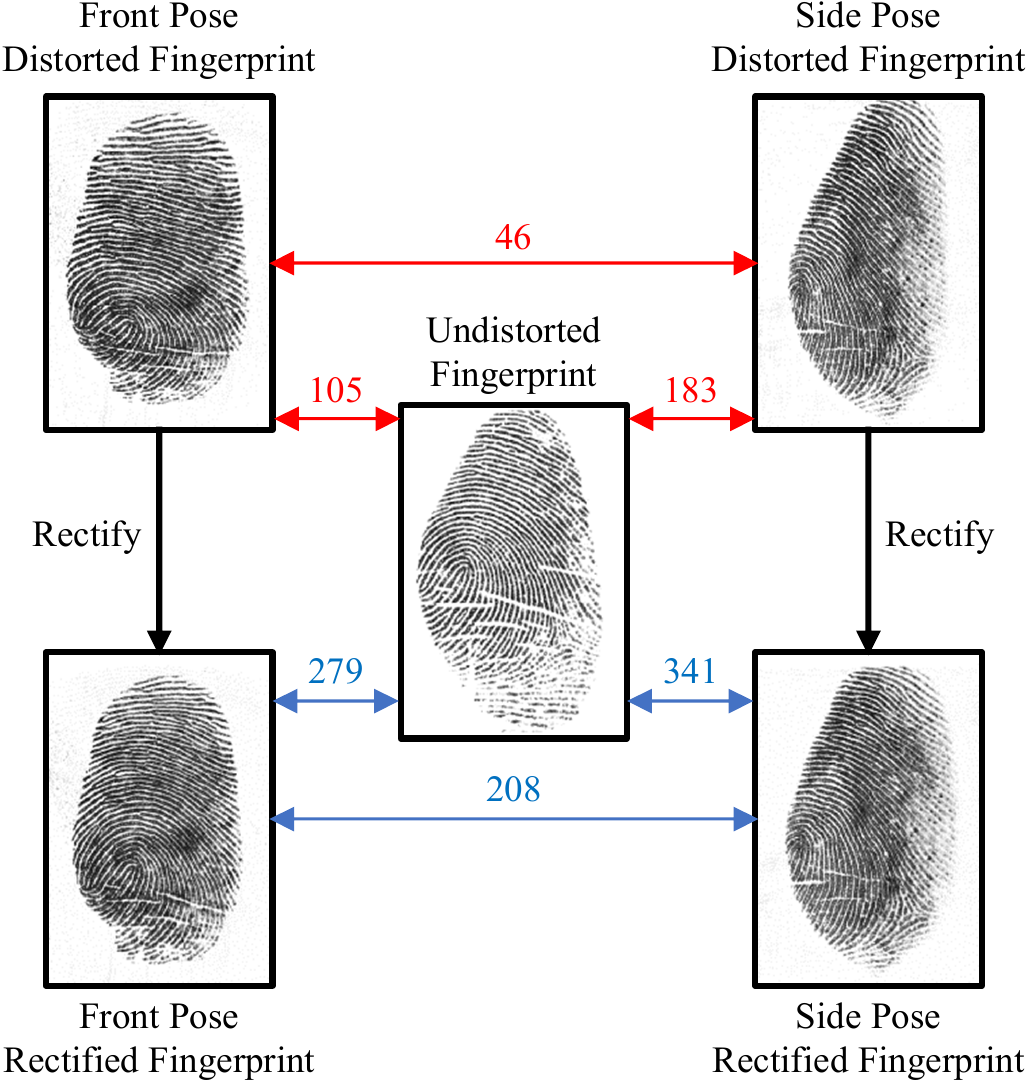}
\caption{Impact of the proposed rectification algorithm on fingerprint matching scores. The red and blue numbers over arrows represent the matching scores before and after rectification calculated by VeriFinger SDK 12.0 \cite{VeriFinger}. Both distorted-undistorted and distorted-distorted matching scores are significantly improved after rectification (105 $\rightarrow$ 279, 183 $\rightarrow$ 341, 46 $\rightarrow$ 208).
}
\label{fig:intro}
\end{figure} 

Since the finger pulp is curved and soft, fingerprint will deform when it is in contact with acquisition equipment, and the distortion field becomes obvious when it is subjected to lateral force or torque \cite{maceo2009qualitative}. In addition, the distortion field of the same finger is different under different pressing poses and strengths, which makes the recognition of distorted fingerprints more challenging. 
Fig. \ref{fig:intro} shows an example, where texture of the same finger has huge differences between two distorted fingerprints and a normal (or undistorted) fingerprint.
In general, fingerprint distortion changes the ridge orientation, ridge frequency, and relative positions of minutiae, which increases intra-class variations among fingerprints from the same finger and thus reduces the matching performance. 
In positive identification scenario such as computer logon, distorted fingerprints will lead to false rejection, very frustrating for the user;
in negative identification scenario, persons in the watch-list may deliberately distort their fingerprints to deceive the recognition system \cite{wein2005using,soweon2012altered},  which is a huge security risk.
Therefore, it is important but still challenging to improve the recognition performance for distorted fingerprints.

\begin{figure}[!t]
	\centering
	\subfloat[]{\includegraphics[width=.95\linewidth]{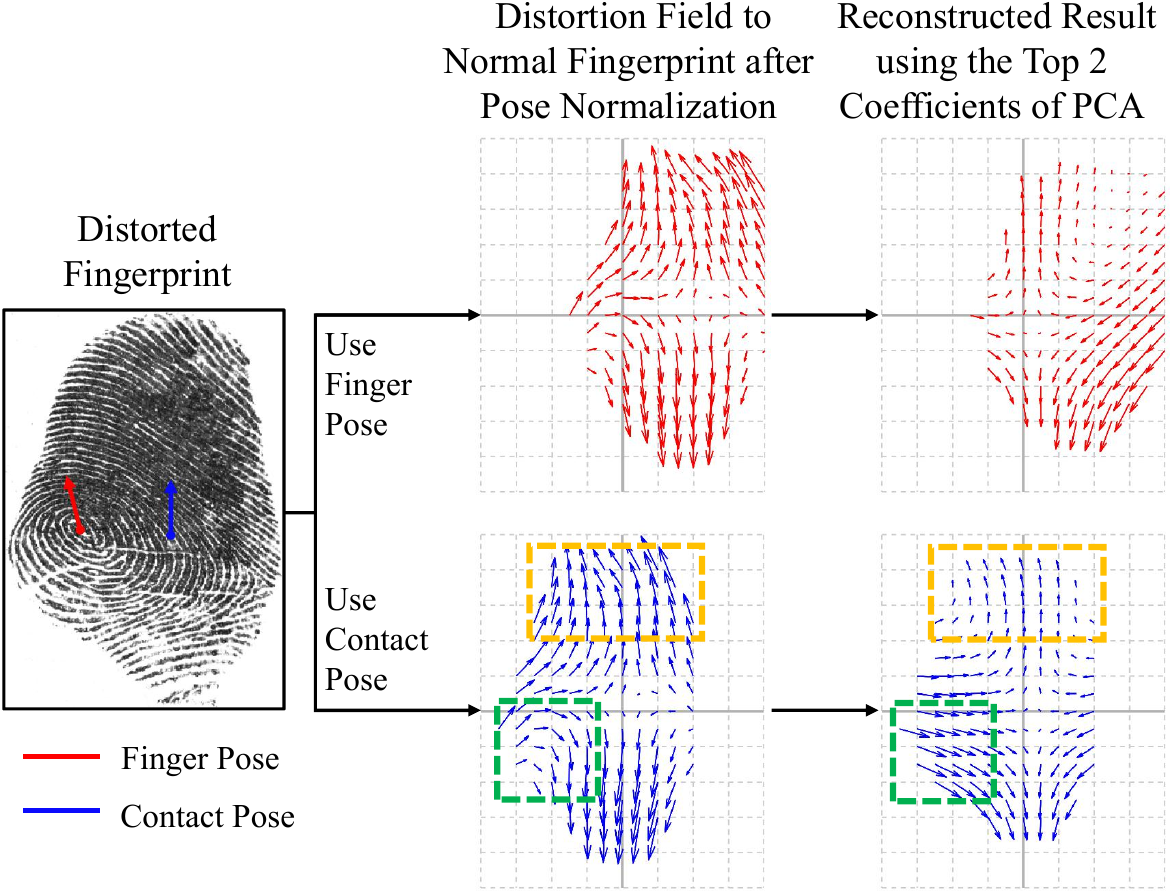}%
		\label{fig:pca_bad_ex}}
	\hfil
	\subfloat[]{\includegraphics[width=.95\linewidth]{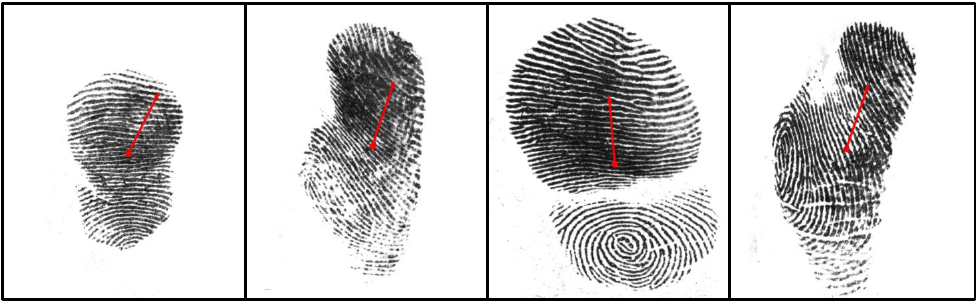}%
		\label{fig:pose_bad_ex}}
	\caption{Failure cases of PCA based rectification. (a) Distortion details are lost. Green and orange rectangles represent regions where orientation or magnitude information is lost. (b) Examples where finger pose is estimated inaccurately. The red arrow indicates the finger pose extimated by \cite{yin2021joint}.}
	\label{fig:pca_pose_bad_ex}
\end{figure}

While given a single distorted fingerprint, distortion rectification algorithms can rectify it to remove distortion components before matching stage, as shown in Fig. \ref{fig:intro}, which is universal (no need to change the existing acquisition and matching modules) and efficient (only performed once after acquisition).
However, conventional rectification algorithms estimate the distortion field represented by Principal Component Analysis (PCA) \cite{si2015detection,gu2018efficient,dabouei2018fingerprint}, whose performance is highly dependent on the performance of pose normalization, and is limited by the accuracy of principle components. 
These methods regard the center of finger pose as the center of principal components when fitting distortion field using PCA, which is inappropriate, because the center of distortion should actually be related to the sliding direction and pressure when the finger presses a contact surface \cite{delhaye2014dynamics}.
Fig. \ref{fig:pca_bad_ex} shows an comparison of the fingerprint distortion field before and after PCA-based reconstruction. 
The definition of finger pose is consistent with the related research \cite{yin2021joint}. 
The center and direction of contact pose are the geometric center and sliding direction of the contact area during pressing respectively.
Obviously, the reconstruction error of PCA is smaller under the contact pose compared to the finger pose.
This example confirms that center of finger pose and distortion are not necessarily consistent, and local details are lost even if pose normalization is accurate.
On the other hand, finger pose itself may be misestimated when dealing with side poses or severely distorted fingerprints, such as examples in Fig. \ref{fig:pose_bad_ex}, which will lead to larger PCA reconstruction errors.

\begin{figure*}[!t]
	\centering
	\includegraphics[width=.95\linewidth]{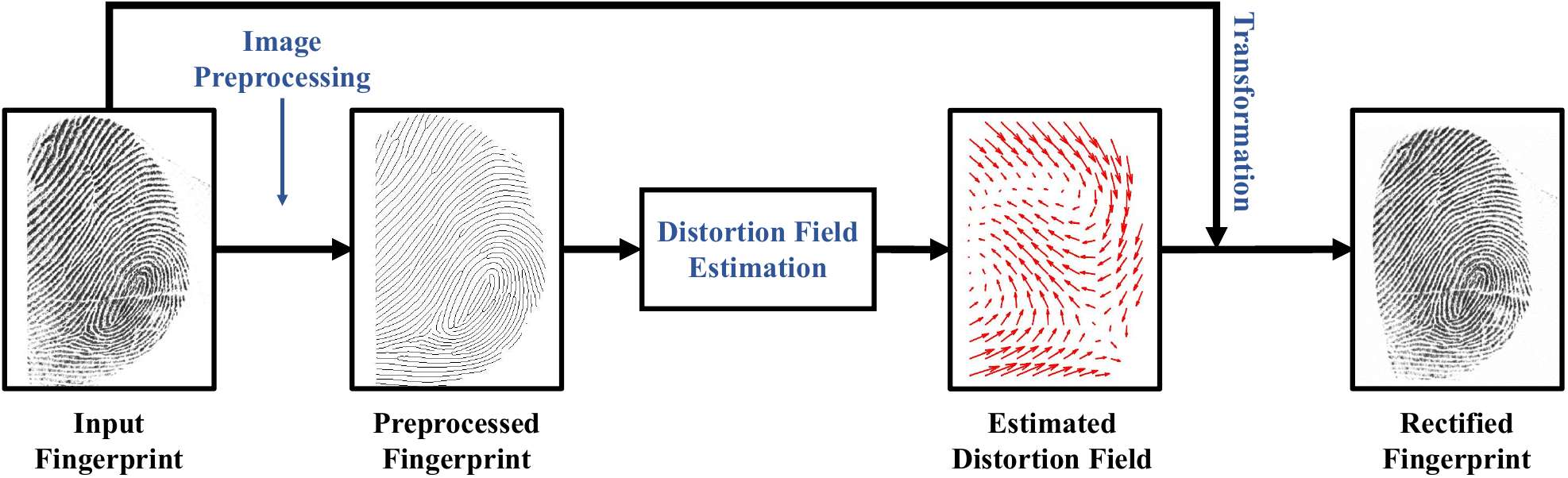}
	\caption{\centering{Flowchart of the proposed distorted fingerprint rectification system.}}
	\label{fig:flowchart}
\end{figure*} 

In this paper, we propose a rectification method, which directly regresses the dense distortion field from a single distorted fingerprint image. Fig. \ref{fig:flowchart} shows the flowchart of the whole rectification algorithm. An input fingerprint is first enhanced and centralized to reduce the impact of image quality, then fed into the proposed network to directly estimate a dense distortion field, and finally transformed into a rectified fingerprint.
The proposed network fused multiple features and captured correlation between local and global distortion patterns so as to finely estimate detailed distortion field, rather than a linear superposition of principle components as previous studies, which can only restore a rough distortion field. Meanwhile, the dependence on finger pose is avoided.

We conducted experiments on FVC2004\_DB1\_A \cite{maio2004fvc2004}, Tsinghua Distorted Fingerprint Database (TDF) \cite{si2015detection}, and a private latent fingerprint database (Hisign Latent Database). Considering the limited number and variety of distorted fingerprints in the existing public dataset TDF, we collected more distorted fingerprints with diverse finger poses and distortion patterns called TDF\_extra, and merged it with TDF as a new database TDF\_V2. 
Two minutiae based matching methods (VeriFinger \cite{VeriFinger} and MCC \cite{cappelli2010minutia}) and two fixed-length descriptor based matching methods (DeepPrint \cite{engelsma2021learning} and an in-house fixed-length descriptor) are used to evaluate rectification accuracy and matching performances.
Experimental results on these datasets show that complicated distorted fingerprints can be rectified and the proposed algorithm achieves the state-of-the-art distortion rectification performance.

This paper is an extension of our conference paper \cite{guan2022direct}, where an orientation feature branch is added and the overall structure is adjusted to help the network better understand fingerprint distortion. Furthermore, a large-scale distorted fingerprint database is synthesized and used for training, instead of real data whose number and variety is limited, to fully utilize the learning capability of neural networks. Experiment databases and evaluation items are also expanded.
Briefly, the contributions of this paper can be summarized as:
\begin{itemize}
	\item We proposed an end-to-end network to directly estimate a dense distortion field without fingerprint pose alignment, instead of its low dimensional representation, from a single fingerprint;
	
	\item Extensive experiments are conducted on three distorted fingerprint databases, which contain normal and distorted fingerprints from optical fingerprint sensor or crime scene. Four fingerprint matching methods have been adopted in the matching experiment, which detailedly shows how fingerprint distortion affects different types of matching methods, at the same time strongly demonstrates the advantages of the proposed method.

\end{itemize}

The rest of this paper is organized as follows.
Section \ref{sec:related works} reviews related works. 
Section \ref{sec:method} introduces the proposed approach. 
Section \ref{sec:data_processing} present the details of training data processing.
Section \ref{sec:experiments} describes the experimental results. 
Finally, Section \ref{sec:conclusion} draws the conclusion.

\section{Related Works}\label{sec:related works}
Researchers have proposed several methods to overcome the negative effects introduced by nonlinear fingerprint distortion, which can be classified into three categories according to in which stage it is performed in the fingerprint recognition systems, namely acquisition, matching and feature extraction stage\cite{si2015detection}.

\subsection{Avoid Distortion in Acquisition Stage}
Some early studies using additional sensors to detect and reject distorted fingerprints in the acquisition stage \cite{ratha1998effect,bolle2000system,dorai2004dynamic,fujii2010detection}, which is expensive and cannot be applied to existing fingerprint databases. 
Some researchers consider contactless 3D fingerprints to avoid skin distortion \cite{Kumar2018,Yin2021,grosz2022c2cl}. However, this technology has not been widely used in real applications. Furthermore, contactless fingerprints usually have perspective distortion, and required strict acquisition conditions to ensure image quality.

\subsection{Tolerate Distortion in Matching Stage}
In order to recognize distorted fingerprints in the matching stage, one way is to appropriately relax the geometrical constraint  \cite{kovacs2000fingerprint,watson2003correlation,jea2005minutia,chen2006new,zheng2007robust,tong2008local,cappelli2010minutia,feng2006fingerprint,lawrence1999systems,engelsma2021learning,takahashi2020fingerprint}. 
Most fingerprint matchers \cite{kovacs2000fingerprint,watson2003correlation,jea2005minutia,chen2006new,zheng2007robust,tong2008local,cappelli2010minutia} are based on minutiae, which model the transformation locally or globally and add distortion tolerance in different ways.
Some fingerprint matchers based on ridge skeleton \cite{feng2006fingerprint} or image \cite{lawrence1999systems} also have similar distortion tolerance strategies. 
In addition, some researchers \cite{engelsma2021learning,takahashi2020fingerprint} use networks to learn fixed-length descriptors of fingerprints, which implicitly increases the tolerance of fingerprint distortion by learning to recognize multiple samples from the same finger.
However, the false match rate will be inevitably increased while allowing large distortion.
Another way is to perform elastic registration of two fingerprints before calculating matching scores \cite{almansa2000fingerprint,bazen2003fingerprint,si2017dense,cui2021dense}. However, elastic registration is very time-consuming for identification systems as it needs to be performed for each matching pair. 

\subsection{Rectify Distortion in Feature Extraction Stage}
Ross \etal \cite{ross2005deformable,ross2006fingerprint} proposed a rectification method based on average deformation pattern of the same finger. However, it is
not always practical because multiple images of the same finger need to be acquired in advance, and it is hard to implement in identification systems.
Moreover, various distortions may still not be covered despite multiple images per finger are available.

Senior and Bolle \cite{senior2001improved} made the first attempt to rectify distortion for a single fingerprint. They assumed that fingerprints have a consistent ridge period and proposed an algorithm to normalize it.
In fact, the ridge period is different even in different regions of the same finger.
Simply normalizing them will lose some important recognition information and even generate more distortion. 

Si \etal \cite{si2015detection} conducted a further study of distortion rectification of a single fingerprint.
They collected the TDF database, which contains 320 videos of distorted fingerprints, and extracted the main distortion patterns by PCA.
They constructed a dictionary of generated distorted fingerprints in the off-line stage and estimated the distortion field of an input fingerprint by nearest neighbor search based on ridge orientation and period features.
Gu \etal \cite{gu2018efficient} improved the nearest neighbor search step in \cite{si2015detection} by utilizing support vector regression to predict the distortion parameters, which greatly improved the efficiency.
Dabouei \etal \cite{dabouei2018fingerprint} used a deep convolution neural network to directly estimate the coefficients of principal components from an input fingerprint. Experiments show that these methods can improve the matching accuracy of most distorted fingerprints in several datasets, but there are still some limitations: (1) normalizing the location and direction of fingerprint in advance is required, which making these algorithms fail to deal with fingerprints whose poses cannot be accurately estimated, such as those with small areas, low quality, or large roll or pitch angles; (2) only principal distortion is estimated, thus distorted fingerprints with complex distortion patterns cannot be accurately rectified. 

\section{Method}\label{sec:method}
\begin{figure*}[!t]
	\centering
	\includegraphics[width=.95\linewidth]{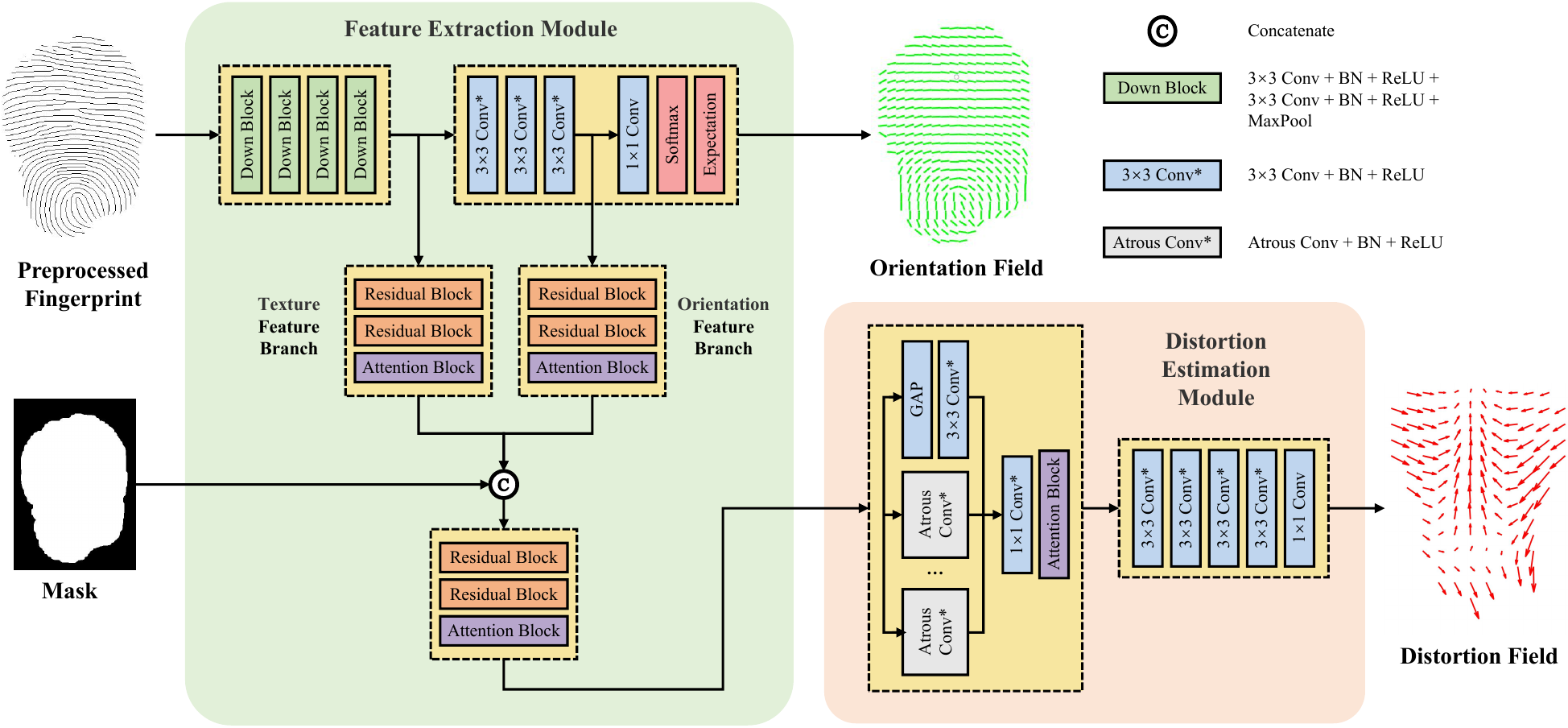}
	\caption{Structure of the proposed distortion estimation network. The network includes a texture branch and an orientation branch for feature extraction, a spatial pyramid block for fusing multi-scale feature information, and a convolution regression block for predicting dense distortion field. For an input distorted fingerprint, the network takes the $16 \times 16$ pixel block as a unit and gives its corresponding distortion field (displacement from the input image to the rectification target).}
	\label{fig:network}
\end{figure*}

In this paper, we aim to predict dense distortion field directly from a single fingerprint using multi-type and multi-scale feature information. 
The complete flowchart of our rectification algorithm is shown in Fig. \ref{fig:flowchart}. For an input fingerprint, our algorithm first obtains its preprocessed image and centers it during preprocessing. 
Then the preprocessed result and its mask are fed into the proposed network, whose structure is shown in Fig. \ref{fig:network}, so as to get the two-dimensional distortion field to its rectification target.
Finally, the rectified fingerprint is obtained by shifting the input image pixel by pixel according to the network prediction result.

\subsection{Image Preprocessing}

\begin{figure}[!t]
	\centering
	\subfloat[]{\includegraphics[height=.65\linewidth]{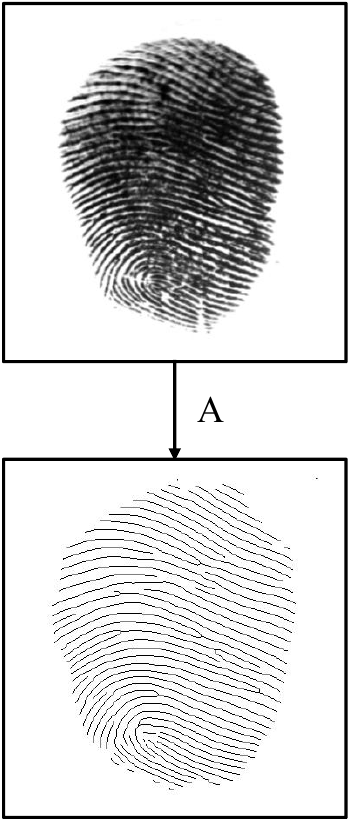}%
		\label{fig:enh_ex_1}}
	\hfil
	\subfloat[]{\includegraphics[height=.65\linewidth]{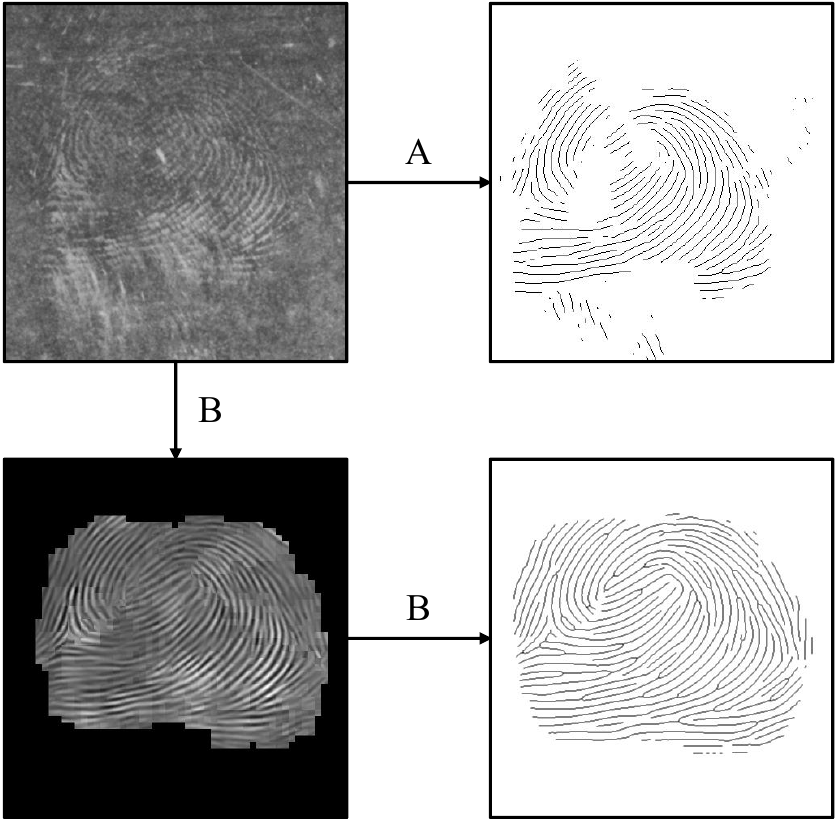}%
		\label{fig:enh_ex_2}}
	\caption{Examples of fingerprint preprocessing. (a) and (b) are respectively collected from the optical sensor and cirme scene. A and B represent different preprocessing strategies.}
	\label{fig:enh_ex}
\end{figure}

Fingerprint distortion usually occurs during the relative sliding process between finger and contact surface, which may lead to texture blurring in some areas and hinder the feature extraction. In addition, the complex background or noise of some latent fingerprints is also a huge challenge.
Therefore, we added the preprocessing step and fed preprocessed results into the distortion regression network, instead of raw image, to minimize the interference of various factors.

Considering that fingerprint distortion is mainly reflected in correlations between ridge regions, skeleton image is taken as the target of preprocessing, which removes the background noise while retaining most of the ridge feature. 
In this preprocessing mode, some raw information, such as ridge width and image grayscale information, may be lost or changed.
However, it has little effect on the judgment of distortion because distortion is not that related to these patterns. 
The impact of preprocessing modes (with varying degrees of change to the raw information) on network performance is compared in subsequent experiments.

Fig. \ref{fig:enh_ex} gives two examples of typical scenarios and corresponding preprocessing strategies. 
For clean fingerprints like Fig. \ref{fig:enh_ex_1}, traditional ridge enhancement and thinning algorithms can be directly applied (VeriFinger \cite{VeriFinger} is used in this paper).
For latent fingerprints, directly extracting the skeleton usually leads to defective effects, as shown in process A of Fig. \ref{fig:enh_ex_2}, whose result has obvious incomplete areas and noisy areas.
In strategy B, we first enhance fingerprint using deep network (FingerNet \cite{tang2017fingernet} is used in this paper), and then extract the skeleton using VeriFinger.
In the example shown in Fig. \ref{fig:enh_ex_2}, strategy B is significantly better than A.
Since directly extracting skeletons from clean fingerprints, i.e. strategy A, has been able to obtain satisfactory results, we only use strategy B on latent fingerprints.
The preprocessed image is then centered so that the network can better focus on features of effective region.
It should be noted that the preprocessed results are only used for regressing distortion field, and the original image is used for rectification, which is centered by the same displacement of its preprocessed image in advance.

\subsection{Network Structure}

Previous rectification methods \cite{si2015detection,gu2018efficient,dabouei2018fingerprint} have proved that extracting and learning distortion patterns from fingerprint video is feasible and effective.
In addition, the dense fingerprint registration method proposed by Cui \etal \cite{cui2021dense} also shows good results in reducing the distortion between paired fingerprints, which predicts the dense distortion field by a siamese block and an encoder-decoder.
Inspired by these methods, although there are no paired input fingerprints in single fingerprint rectification, we still construct the reference relationship among the features on different spatial scales, so as to learn the patterns of dense distortion field. 

Fig. \ref{fig:network} illustrates the structure of our dense distortion regression network.
Preprocessd fingerprint is used for input because we expect the proposed network to focus only on distortion estimation so as to avoid unnecessary tasks making it redundant.
Besides, the network also takes fingerprint mask as input due to a mask-related constraint is used when generating the ground truth. Meanwhile, it is also used to indicate the effective finger region in loss functions.
Mask here is the largest connected domain of the segmentation result, which is obtained by calculating the gradient and setting a threshold for the input preprocessed image. 

\begin{figure}[!t]
	\centering
	\includegraphics[width=.98\linewidth]{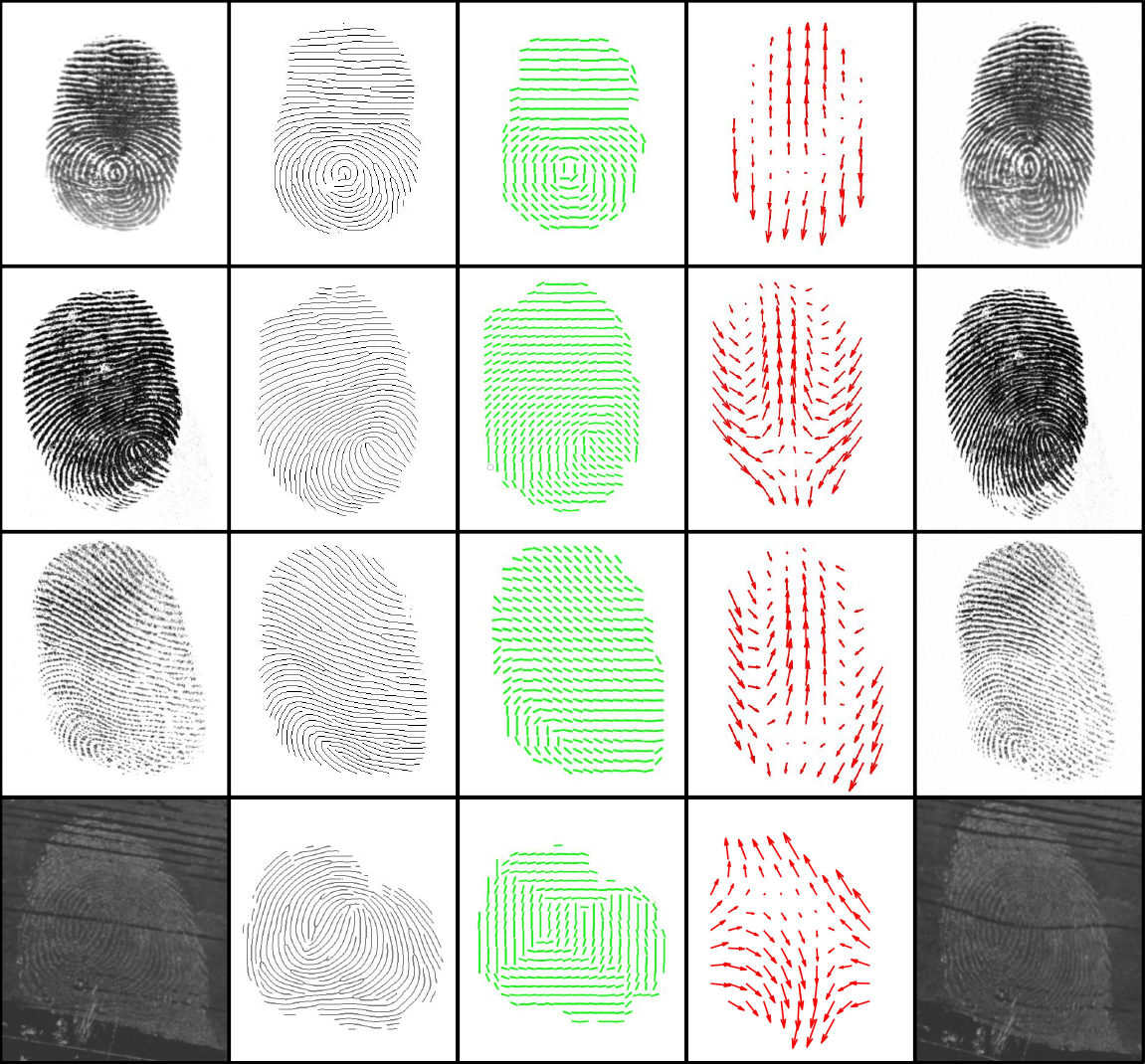}%
	\caption{Input, output, and intermediate results in distortion estimation and rectification. Columns from left to right corresponds to original fingerprints, preprocessed fingerprints, orientation field and distortion field estimated by the proposed network, and their rectified results.}
	\label{fig:network_ex}
\end{figure}

The network can be divided into two parts. 
In the feature extraction module, texture features are first obtained through $4$ downsampling blocks.
Furthermore, considering that fingerprint distortion changes the arrangement of ridges in some regions, orientation feature is extracted by concatenating $3$ convolution layers and a regression layer, which can well reflect this local anomaly (see Fig. \ref{fig:network_ex}, in which the transition of orientation field in distorted regions is obviously abrupt).
The orientation output is classification probabilities mapped by softmax activation, which has $180$ categories evenly divided from $-90^{\circ}$ to $90^{\circ}$ and $1/16$ size of inputs.
Features are further refined through a sub-module consisting of $2$ residual blocks, which is used to improve the ability to express features while avoiding gradient problems caused by too deep layers, and a channel attention block \cite{hou2021coordinate}, which is coordinate-sensitive to help the model locate and identify interest regions.
Features of two branches and the $1/16$ scaled mask are concatenated and then fused through the refinement sub-module mentioned above. 
It should be noted that we add the mask after branch feature extraction instead of before, because it is used to constrain the estimation of distortion and not helpful for analyzing the regional features.

In the distortion estimation module, a dense distortion field from the input fingerprint to the rectification target is directly regressed according to the mixed features.
Since fingers have a continuous irregular elastic surface, the distortion field of the distorted fingerprint is usually relatively uniform in a large area, and it is locally different at the same time.
In order to capture those contextual information at multiple scales, global average pooling and atrous convolution blocks (with dilated ratio of $1$, $2$, $4$, $8$) with different dilation rates are parallelized in the spatial pyramid module, inspired by \cite{chen2018encoder}.  
In the last part, several convolution blocks are used to regress the distortion field.
The output regards $16\times16$ pixel block as a displacement unit, which is fine enough to correct the distortion of the fingerprint (similarly, orientation field is usually scaled to $1/8$ size of the original image in the orientation estimation task) and make the network lightweight at the same time. The estimated distortion field is interpolated to the original size and then used in the transformation step.
Examples of network inputs and outputs are shown in Fig. \ref{fig:network_ex}.

\subsection{Loss Function}
The total loss $\mathcal{L}$ is a weighted sum of two types of losses, which corresponds to the network outputs, namely orientation loss $\mathcal{L}_{\rm{ori}}$ and distortion loss $\mathcal{L}_{\rm{dis}}$.

\subsubsection{Orientation Loss}
Ridge orientation in local patches is relatively similar, which means an unbalanced multi-class distribution. Therefore, focal loss is applied as it is proved efficacious for this problem. The orientation classification loss $\mathcal{L}^{\rm{cla}}_{\rm{ori}}$ is defined as
\begin{equation}
	\begin{aligned}
		\mathcal{L}^{\rm{cla}}_{\rm{ori}} & = - \frac{1}{|M|} \sum_{M} \sum_{t=1}^T \alpha\left(1-q^t\right)^\gamma \log \left(q^t\right) , \\
		q^t & =y^t p^t+\left(1-y^t\right)\left(1-p^t\right) .
		\label{eq:ori_cla}
	\end{aligned}
\end{equation}
where $M$ is the input mask, $p^t$ and $y^t$  are the probability of the $t$-th category according to estimation results and ground truth respectively. $T$ is the total number of orientation categories which is set to $180$ in the experiment. $\alpha$ and $\gamma$ are hyper-parameters of focal loss.

Furthermore, orientation coherence, a strong domain prior knowledge \cite{kass1987analyzing}, is introduced as $\mathcal{L}^{\rm{smo}}_{\rm{ori}}$ to constrain the smoothness of estimated orientation: 
\begin{equation}
	\begin{aligned}
	\mathcal{L}^{\rm{smo}}_{\rm{ori}}&=\frac{|M|}{\sum_{M} Coh}-1,\\
	Coh &= \sum_M \frac{\left\|\left(\bar{d}_{\cos}(x, y), \bar{d}_{\sin }(x, y)\right) \ast \rm{K}\right\| } {\bar{d}(x, y)\ast \rm{K}},
	\end{aligned}
	\label{eq:loss_ori_smo} 
\end{equation}
where $\rm{K}$ is a $3 \times 3$ all-one kernel and $M$ is the input mask.
Let the definitions of $T$ and $p^t$ in consistent with Equation \ref{eq:ori_cla}, values $\bar{d}_{\cos }$, $\bar{d}_{\sin }$ and $\bar{d}$ at point $(x,y)$ are computed as
\begin{equation}
	\begin{aligned}
		& \bar{d}_{\cos }(x, y)=\frac{1}{T} \sum_{t=1}^{T} p^{t}(x, y)\cos \left(\frac{360}{T}t\right), \\
		& \bar{d}_{\sin }(x, y)=\frac{1}{T} \sum_{t=1}^{T} p^{t}(x, y)\sin \left(\frac{360}{T}t\right), \\
		& \bar{d}(x, y) = \left\|(\bar{d}_{\cos}(x, y), \bar{d}_{\sin}(x, y) )\right\|.
	\end{aligned}
\end{equation}

\begin{figure*}[!t]
	\centering
	\subfloat[]{\includegraphics[width=.6\linewidth]{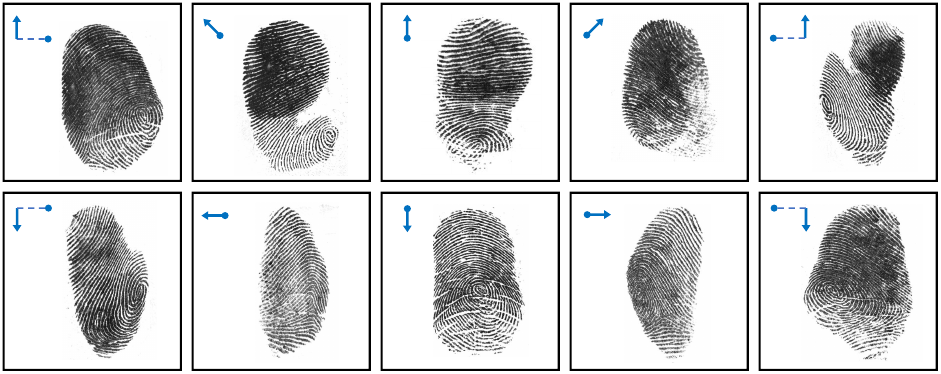}%
	}
	\hfil
	\subfloat[]{\includegraphics[width=.37\linewidth]{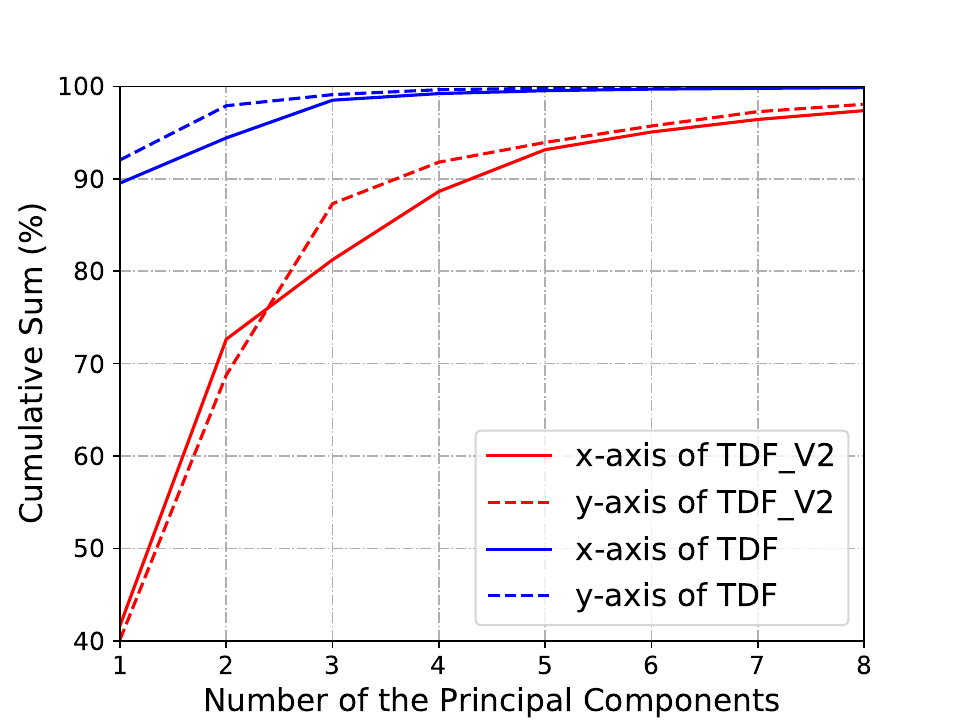}%
		\label{fig:new_db_cumsum}}
	\caption{Statistics of the new distorted fingerprint database TDF\_V2. (a) Examples of all 10 distortion types in the additional data TDF\_extra. The solid blue circle represents finger of front pose, the arrow represents the direction of force when pressing, and the dotted line represents finger of side pose. (b) Cumulative sum of the main distortion patterns in TDF and TDF\_V2. The numbers on abscissa represent descending ranks of the top eight principal components in two axes. Finger pose of each fingerprint is normalized in advance.}
	\label{fig:new_db}
\end{figure*}

\subsubsection{Distortion Loss}
It consists of two parts, namely the regression loss $\mathcal{L}^{\rm{reg}}_{\rm{dis}}$ between the estimated distortion field $F^{{\rm est}}$ and the ground truth $F^{{\rm gt}}$, and the smoothing loss $\mathcal{L}^{\rm{smo}}_{\rm{dis}}$ according to the gradient of $F^{{\rm est}}$:
\begin{align}
	\mathcal{L}^{\rm{reg}}_{\rm{dis}}&=\frac{1}{|M|}  \sum_{M} \left\| F^{{\rm 	est}}-F^{{\rm gt}}\right\|^{2}, \label{eq:loss_dis_reg} \\
	\mathcal{L}^{\rm{smo}}_{\rm{dis}}&= \frac{1}{|M|} \sum_{M}\left(  \left\|\nabla F_x^{\rm est }\right\|^{2}+\left\|\nabla F_y^{\rm est}\right\|^{2}\right), \label{eq:loss_dis_smo}
\end{align}
where $F_x^{\rm est }$ and $F_y^{\rm est }$ are components of $F^{\rm est}$ in the corresponding dimension. $\nabla$ represents the spatial gradient operation in both two dimensions.

\subsubsection{Final Loss}
Loss sub-functions described above are weighted and summed as the final loss, which is
\begin{equation}
	\mathcal{L} = \lambda_{{\rm ori}}(\mathcal{L}^{\rm{cla}}_{\rm{ori}}+w_{\rm{ori}} \cdot \mathcal{L}^{\rm{smo}}_{\rm{ori}})+\lambda_{{\rm dis}}(\mathcal{L}^{\rm{reg}}_{\rm{dis}}+w_{\rm{dis}} \cdot \mathcal{L}^{\rm{smo}}_{\rm{dis}}).
\end{equation}
All hyper-parameters are determined empirically and listed in Table \ref{tab:parameters}.

\begin{table}[!t]
\caption{Hyper-Parameters used in Loss Function\label{tab:parameters}}
\centering
\renewcommand{\arraystretch}{1.4}
\begin{tabular}{|c|c c c c c c|}
	\hline
	Parameter &$\alpha$ & $\gamma$ & $\lambda_{{\rm ori}}$ & $\lambda_{{\rm dis}}$ & $w_{\rm{ori}}$ & $w_{\rm{dis}}$\\
	\hline
	Value &1.0 & 2.0 & 1.0 & 1.0 & 0.5 & 1.0\\
	\hline
\end{tabular}
\renewcommand{\arraystretch}{1}
\end{table}

\subsection{Image Transformation}
The final transformation result is a nearest neighbor interpolation performed on the original image centered in preprocessing stage according to the regressed distortion field $F^{{\rm est}}$.
It is a dense displacement so that the pixel-by-pixel correspondence can be directly obtained, rather than a rough trend computed by PCA in previous methods \cite{si2015detection,gu2018efficient,dabouei2018fingerprint}. 
Hence local distortion can be rectified more freely and finely.
Since the smoothing loss (Equation \ref{eq:loss_ori_smo} and \ref {eq:loss_dis_smo}) are used in network training, the spatial continuity of the predicted value can be guaranteed.
From the comparison of the first (original image) and last column (transformed image) in Fig. \ref{fig:network_ex}, it can be seen that fingerprint distortion is significantly reduced by the proposed rectification algorithm.

\section{Generation of Training Data}\label{sec:data_processing}
In this section, preprocessing details of generating training data are introduced. 
Public dataset TDF with additional data is used for generating distortion fields and then applied to normal fingerprints with multiple poses.
In this way, a large-scale synthetic dataset is synthesized for training and validation, so that the learning capability of neural networks can be fully utilized.

\subsection{Generation of Distortion Fields} \label{sec:distortion_fields_generation}
There is still a lack of systematic theory about the fingerprints distortion law at present, and consequently a large amount of multi-type distorted fingerprints are required to cover the practical scenarios as much as possible. 
However, it is very difficult and time-consuming to directly collect a large number of diverse distorted fingerprints and obtain their accurate distortion fields.
Therefore, we model geometric deformations on real dataset of limited size and generate synthetic distortion fields according to it.
Distorted fingerprints are then generated conveniently and efficiently by applying synthetic distortion fields on normal fingerprints, in which the quantity and variety of distortion type, distortion strength, fingerprint pattern, and finger pose can be guaranteed.
Meanwhile, the ground truth of distortion field is completely accurate in this way.

Existing TDF database established by Si \etal \cite{si2015detection} has 320 distorted fingerprints, whose size is relatively small.
Accordingly, we collected more abundant distorted fingerprints called TDF\_extra in addition to it, and merged them as a new database named TDF\_V2 for convenience. The images are cropped to $512\times512$, and the finger pose of each fingerprint is normalized using the method in \cite{yin2021joint}.
In the collection process of TDF\_extra, fingers were first pressed in the front pose, and then deformed by horizontal or vertical force or torque. 
Considering that the common fingerprint distortion in practice is usually caused by unidirectional force and may occur with various poses, we collected additional 480 distorted fingerprint videos generated by one-way rubbing in different directions on the front or side finger pose, instead of torque and two-stage force types in TDF. 
A Frustrated Total Internal Reflection (FRIR) fingerprint scanner was used to obtain fingerprint sequences. 
Totally $10$ distortion types of videos for each finger are sampled at $30$Hz with a resolution of $500$ ppi, and the duration of each video is about $3$s. 
Examples of 10 distortion fingerprint types in the additional data and principal component distributions before and after mergence are shown in Fig. \ref{fig:new_db}. It can be seen that with different initial poses and pressing directions, different distortion patterns will be produced (it takes about 6-8 principal components to represent the distortion field well in TDF\_V2, instead of 2 for TDF).

Similar to \cite{si2015detection}, we take the initial frame as the normal fingerprint and the end frame as the distorted fingerprint, and obtain the displacement between them by pairing minutiae points in adjacent frames and performing thin-plate spline interpolation. 
Geometric distortion is then modeled by PCA using $400$ samples in TDF\_V2.
The generation function of synthetic distortion field $F$ (from the distorted fingerprint to its rectification target) can be expressed as
\begin{equation}
	F = F_{0} + \sum_{i=1}^{t} c_i \sqrt{\lambda_{i}} E_{i},
\end{equation}
where $F_{0}$ is the mean distortion field, $\lambda_{i}$ and $E_{i}$ represent the eigenvalue and eigenvector of the $i$-th principal component. 
Since the distortion in this dataset is more complex than TDF (as shown in Fig. \ref{fig:new_db_cumsum}), we use the top $8$ principal components instead of $2$ in \cite{si2015detection,gu2018efficient,dabouei2018fingerprint}, that is, $t=8$. Eigenvector component coefficient $c_i$ is randomly select from $-2.0$ to $2.0$.

\begin{figure}[!t]
	\centering
	\includegraphics[width=.95\linewidth]{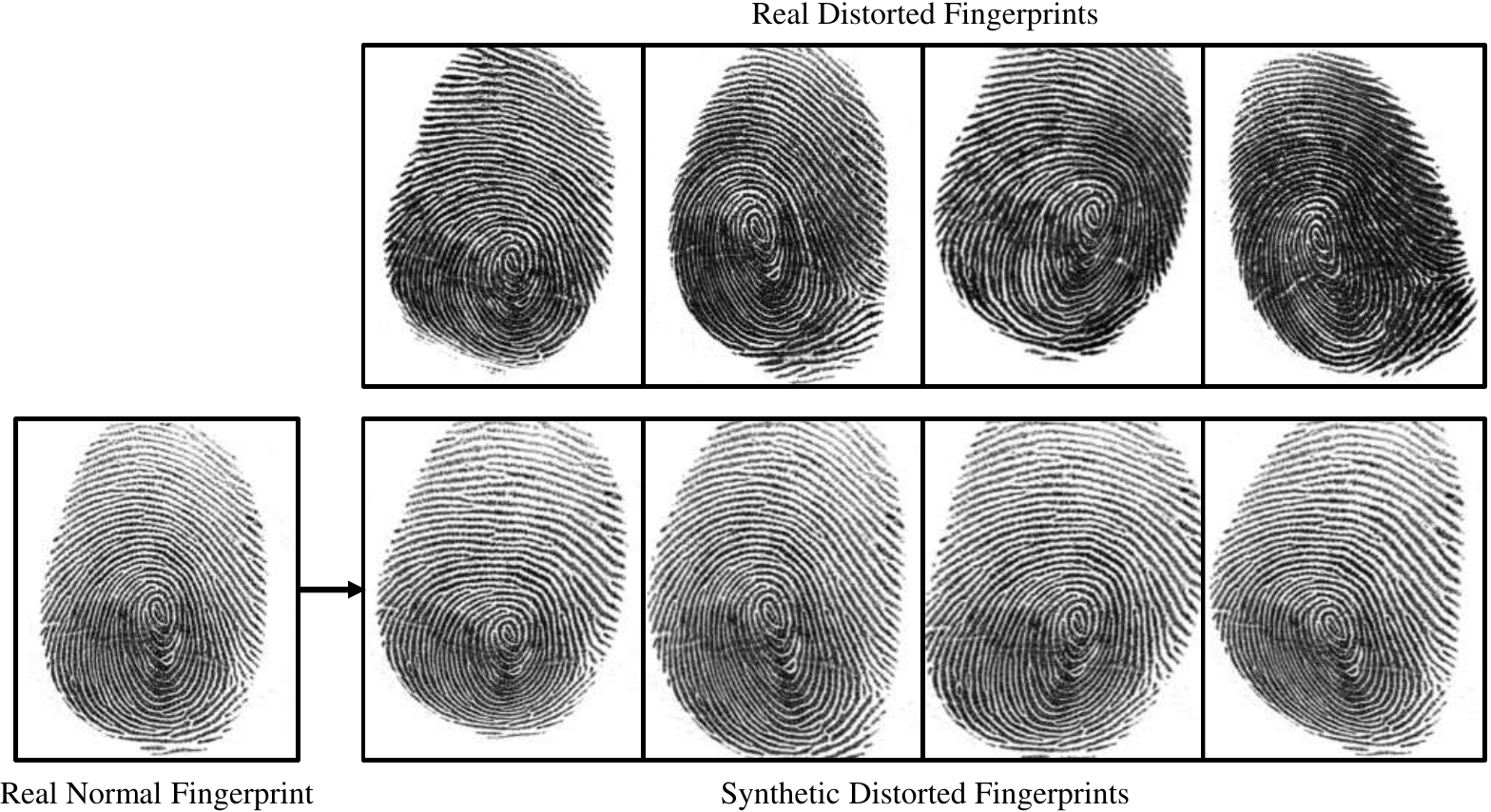}
	\caption{Examples of distorted fingerprint generation. The upper and lower images of each column correspond to real and synthetic fingerprints with approximate distorted patterns. Synthetic fingerprints are generated using the same real normal fingerprint. All fingerprints correspond to the same finger.}
	\label{fig:fp_generate_ex}
\end{figure}

\subsection{Training Data Building} \label{sec:training_data_building}

The Diverse Pose Fingerprint Database is used for generation as the normal fingerprint, which records the sequence frames when participants roll their fingers. In this way, various poses can be obtained conveniently, and the fingerprints of each frame are almost undistorted with the cooperation of volunteers. 
The entire database contains $686$ fingers, of which $558$ are used for training and the rest for validation. 
There is no overlapping data in these two subsets.
We randomly pick $5$ images from each finger sequence and then mirror them to further augment fingerprint types.
To enhance the robustness of the input pose, each fingerprint is rotated by $90$, $180$ and $270$ degrees. It should be emphasized that these rotation augmented parts are not used when reimplementing the PCA-based methods \cite{si2015detection,gu2018efficient,dabouei2018fingerprint} since they normalize fingerprints according to their poses.

Denote subscript $1$ and $2$ as normal fingerprints and generated distorted fingerprints, their relationship with synthesized distortion field $F_{21}$ is
\begin{align}
	I_{2}(x+dx_{12},y+dy_{12}) &= I_1(x,y),\\
	M_{2}(x+dx_{12},y+dy_{12}) &= M_1(x,y),\\
	F_{12}(x+dx_{21},y+dy_{21}) &= -F_{21}(x,y),
\end{align}
where $I$ and $M$ are the fingerprint image and mask, $dx_{ij}$ and $dy_{ij}$ are the dimensional components of $F_{ij}$ at point $(x, y)$.
Non DC (Direct Current) components in $F_{21}$ are extracted as follows:
\begin{equation}
	\begin{aligned}
		\hat{F}_{{\rm 21}}(x,y)&=\left(R_{{\rm 12}} \cdot P_{{\rm 1}}(x,y)+t_{{\rm 12}}\right)-P_{{\rm 2}}(x,y) ,\\
		R_{{\rm 12}}, t_{{\rm 12}}&=\underset{R, t}{\arg \min }\frac{1}{|M_2|} \sum_{M_2}\left\|\left(R \cdot P_{{\rm 1}}+t\right)-P_{{\rm 2}}\right\|
	\end{aligned}
	\label{eq:constraint}
\end{equation}
in which $P_{i}$ is the paired set of coordinates $1$ and $2$, $R$ and $t$ are rigid rotation and translation matrices.
With the strong constraint proposed above, the situation of multiple solutions due to rigid transformations is avoided, and unnecessary DC component caused by finger translation or rotation during acquisition will not be introduced. In this way, distorted fingerprints $I_2$ and ground truth $\hat{F_{\rm{21}}}$ are generated and used for training.
We randomly synthesize $5$ distorted fingerprints with each normal fingerprint in augmented DPF, that is, the final dataset has $111,800$ training samples and $25,400$ validation samples. 
Examples of distorted fingerprints generation are given in Fig. \ref{fig:fp_generate_ex}, which illustrates that our method of synthesizing distortion field can approximate real scenes.

\section{Experiments}\label{sec:experiments}
In this section, we compare the proposed method with state-of-the-art rectification algorithms in terms of distortion estimation accuracy, rectified fingerprint matching performance, and inference efficiency.
Meanwhile, ablation studies are listed to verify the effectiveness of network sub-modules.

\subsection{Comparison Methods}
Due to the more complicated modeling of geometric distortion (as shown in Fig. \ref{fig:new_db_cumsum}), PCA based methods \cite{gu2018efficient,dabouei2018fingerprint} use the top $8$ principal components instead of $2$ in our reimplementation. 
Furthermore, since the algorithm of Gu \etal \cite{gu2018efficient} is a convex optimization method only suitable for small data sets, $5000$ samples from the synthetic dataset are selected for training.
The method proposed by Si \etal \cite{si2015detection} is not included since the algorithm is highly dependent on the dictionary size and the retrieval process is significantly time-consuming, making it difficult to apply in practice, and previous studies showed that its performance has been exceeded by \cite{gu2018efficient,dabouei2018fingerprint}.
In particular, we compare the performance of proposed method with (complete network which has better performance) or without (basic network which has smaller model size) fusing the orientation feature branch 
to make it suitable for different application scenarios, distinguished by `+O' in the following.

\subsection{Experimental Datasets}
Public databases FVC2004\_DB1\_A \cite{maio2004fvc2004}, expanded Tsinghua Distortied Fingerprint Database \cite{si2015detection} (without data used in Section \ref{sec:distortion_fields_generation}, called TDF\_V2\_T), a latent fingerprint dataset (Hisign Latent Database, called HLD), and a verification database synthesized on sequence fingerprints described in Section \ref{sec:training_data_building} (called DPF\_syn\_valid) are used for experiments. 
In order to evaluate the performance of distortion rectification algorithms more exhaustively, hard subsets are selected from each database according to the genuine match score of original fingerprints calculated by VeriFinger \cite{VeriFinger}.
Table \ref{tab:database} briefly introduces the contents and application details of all databases.
These data include distorted and undistorted fingerprints captured from optical instruments or crime scenes, which are suitable for comprehensive evaluating effects of distortion before and after rectification.
All images are uniformly pose corrected and cropped to $512 \times 512$ before rectification.

\begin{table*}[!t]
	\caption{Fingerprint Databases Used in Experiments\label{tab:database}}
	\centering
	\renewcommand{\arraystretch}{1.4}
	\begin{tabular}{| m{1.5cm}<{\centering} | m{3.7cm}<{\centering} | m{3.7cm}<{\centering} | m{3.7cm}<{\centering} | m{3.7cm}<{\centering} |}
		\hline
		\textbf{Database} & FVC2004\_DB1\_A & TDF\_V2\_T & HLD & DPF\_sym\_valid \\
		\hline
		\quad\newline\quad\newline\quad\newline\textbf{Image} 
		& \begin{minipage}[][20mm][c]{.1\textwidth}\centering
			\includegraphics[width=\linewidth]{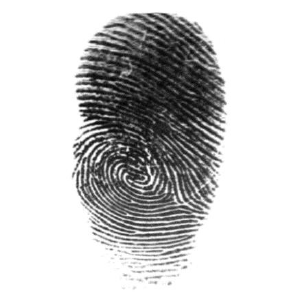}
		\end{minipage}
		& \begin{minipage}[][20mm][c]{.1\textwidth}\centering
		\includegraphics[width=\linewidth]{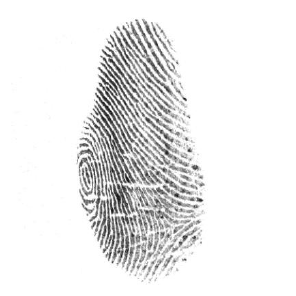}
		\end{minipage}
		& \begin{minipage}[][20mm][c]{.1\textwidth}\centering
			\includegraphics[width=\linewidth]{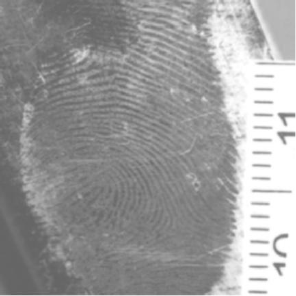}
		\end{minipage}
		& \begin{minipage}[][20mm][c]{.1\textwidth}\centering
			\includegraphics[width=\linewidth]{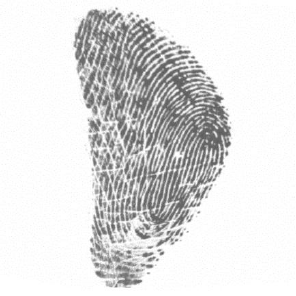}
		\end{minipage} \\
		\hline
		\textbf{Sensor} & Optical & Optical & Inking / Optical, Latent & Synthetic \\
		\hline
		\textbf{Description}
		& \multicolumn{1}{m{3.7cm}|}{$800$ fingerprints from $100$ fingers. Most of the poses are frontal. Some fingerprints have large distortion.} 
		& \multicolumn{1}{m{3.7cm}|}{$400$ pairs of normal and mated highly distorted fingerprints from $153$ fingers. Various poses are included.}
		& \multicolumn{1}{m{3.7cm}|}{$1467$ pairs of rolled / plain and mated latent fingerprints from $1467$ fingers in different distortion levels. Various poses are included.}
		& \multicolumn{1}{m{3.7cm}|}{$25,400$ fingerprints generated from $635$ original images of $127$ fingers. Various poses, distortion types and degrees are included.}\\
		\hline
		\textbf{Experiments} 
		& Distoriton estimation accuracy \newline Matching accuracy
		& Distoriton estimation accuracy \newline Matching accuracy
		& Distoriton estimation accuracy \newline Matching accuracy
		& Ablation study\\
		\hline
		\textbf{Genuine Match} 
		& \multicolumn{1}{m{3.7cm}|}{Fingerprints of the same finger are matched each other while symmetric matches are avoided. $2800$ matches in total and $336$ matches in subset.}
		& \multicolumn{1}{m{3.7cm}|}{Each distorted fingerprint is matched with all mated normal fingerprints from the same finger. $2014$ matches in total and $588$ matches in subset.}
		& \multicolumn{1}{m{3.7cm}|}{Each latent fingerprint is mathched with its mated rolled fingerprint. $1467$ matches in total and $765$ matches in subset.}
		& \multicolumn{1}{m{3.7cm}<{\centering}|}{\textbackslash}\\
		\hline
		\textbf{Impostor Match} 
		& \multicolumn{1}{m{3.7cm}|}{First fingerprints of each finger are matched each other while symmetric matches are avoided. $4950$ matches both in total and hard subset.}
		& \multicolumn{1}{m{3.7cm}|}{Each distorted fingerprint is matched with all normal fingerprints from other fingers, $317,305$ matches in total and $59,412$ matches in subset.}
		& \multicolumn{1}{m{3.7cm}|}{Each latent fingerprint is matched with all non-mated rolled fingerprints, $2,150,622$ matches in total and $1,121,490$ matches in subset.}
		& \multicolumn{1}{m{3.7cm}<{\centering}|}{\textbackslash}\\
		\hline
	\end{tabular}
	\renewcommand{\arraystretch}{1}
\end{table*}

\subsection{Distortion Estimation Accuracy}\label{sec:exprtiment_distortion_acc}
Since there is no ground truth in real datasets, we use the mean reprojection error (MRE) of minutiae between distorted fingerprints and their corresponding normal fingerprints to approximately evaluate the distortion estimation accuracy. 
The average number of matching pairs (AP) is also calculated, because the more accurate the estimation and rectification is, the more minutiae can be paired. 
Verifinger \cite{VeriFinger} is used for extracting and pairing minutiae as it has been proven to be very accurate on this. 
Let $N$ and $m_n$ denote the number of dataset samples and matching minutiae of the $n$-th sample pair,  $P_{{\rm{dis}}}$ and $P_{{\rm{norm}}}$ denote the corresponding points in distorted and mated normal fingerprints, the indicator $\rm{MRE}$ and $\rm{AP}$ is expressed as
\begin{equation}
	\begin{aligned}
		\rm{MRE}&= \frac{1}{N} \sum_{N} {\rm{MRE}}_{n}, \quad {\rm{AP}} = \frac{1}{N} \sum_{N} {m_{n}},\\
		\rm{MRE}_{n}&= \frac{1}{m_n} \sum_{m_n} \left\| \left(\hat{R}_n \cdot P_{{\rm{dis}}}+\hat{t}_n \right)-P_{{\rm{norm}}} \right \| ,\\
		\hat{R}_n, \hat{t}_n &=\underset{R, t}{\arg \min }\frac{1}{\rm{m_n}} \sum_{m_n}\left\|\left(R \cdot P_{\rm{dis}}+t\right)-P_{{\rm{norm}}}\right\|.
	\end{aligned}
	\label{eq:MRE}
\end{equation}
As shown in Table \ref{tab:MRE}, the proposed method leads prominently in each dataset and only slightly inferior in the average pairing numbers on a latent database subset, which suggests that directly regressing the dense distortion field is more refined. At the same time, it also proves that fusing orientation features can indeed assist the network to better capture distorted information.

Moreover, generated validation set DPF\_sym\_valid is used to verify the value of specific modules, which has accurate distortion field ground truth.
The ablation experiments on distortion estimation accuracy are given in Table \ref{tab:ablation}, where the indicator $\mathcal{L}_{{\rm dis}}^{\rm{reg}}$ is defined in Equation \ref{eq:loss_dis_reg}.
Similarly, the distortion estimation performance under different preprocessing and training strategies is evaluated and listed in Table \ref{tab:ablation_strategies}. Among them, `enhancement', `binarization', and `thinning' respectively indicate that the fingerprint is preprocessed into enhanced image (using Gabor filters which can retain most of the raw information), binary image (lose grayscale intensity), and skeleton image (lose grayscale intensity and ridge width); `single-stage' and `two-stage' respectively indicate that the orientation field and distortion field branches of proposed network are trained together or in two-stage. From the comparison it can be seen that removing some redundant information in the preprocessing stage can help the network better focus on distortion estimation. On the other hand, the performance of proposed network decreases when trained in two-stage, which indicates that estimating distortion field accurately requires more information than the orientation field and joint training helps the network to better extract required features.

\begin{table*}[!t]
	\caption{Distortion Estimation Accuracy of Different Rectification Methods.\label{tab:MRE}}
	\centering
	\renewcommand{\arraystretch}{1.5}
	\begin{tabular}{|m{3.2cm}<{\centering}|m{1.6cm}<{\centering}|m{1.6cm}<{\centering}|m{1.6cm}<{\centering}|m{1.6cm}<{\centering}|m{1.6cm}<{\centering}|m{1.6cm}<{\centering}|}
	\hline
	\multirow{3}{*}{\textbf{Methods}} & \multicolumn{6}{c|}{$\mathbf{MRE\,(\downarrow)}$, $\mathbf{AP\,(\uparrow)}$}\\
	\cline { 2 - 7 } 
	& \multicolumn{2}{m{3.2cm}<{\centering}|}{\textbf{FVC2004\_DB1\_A}\quad\quad\quad\quad}
	& \multicolumn{2}{m{3.2cm}<{\centering}|}{\textbf{TDF\_V2\_T}\quad\quad\quad\quad}  
	& \multicolumn{2}{m{3.2cm}<{\centering}|}{\textbf{HLD}\quad\quad\quad\quad}  \\
	\cline { 2 - 7 }
	& \multicolumn{1}{m{1.6cm}<{\centering}}{\textbf{full}}
	& \multicolumn{1}{m{1.6cm}<{\centering}|}{\textbf{subset}}
	& \multicolumn{1}{m{1.6cm}<{\centering}}{\textbf{full}}
	& \multicolumn{1}{m{1.6cm}<{\centering}|}{\textbf{subset}}  
	& \multicolumn{1}{m{1.6cm}<{\centering}}{\textbf{full}}
	& \multicolumn{1}{m{1.6cm}<{\centering}|}{\textbf{subset}}  \\
	\hline
	\multicolumn{1}{|m{3.2cm}<{\centering}|}{Without Rectification}
	& \multicolumn{1}{m{1.6cm}<{\centering}}{7.66, 32.32}
	& \multicolumn{1}{m{1.6cm}<{\centering}|}{12.39, 23.96}
	& \multicolumn{1}{m{1.6cm}<{\centering}}{12.17, 31.58}
	& \multicolumn{1}{m{1.6cm}<{\centering}|}{12.80, 22.64}  
	& \multicolumn{1}{m{1.6cm}<{\centering}}{13.54, 22.89}
	& \multicolumn{1}{m{1.6cm}<{\centering}|}{15.17, 18.26}  \\
	
	\multicolumn{1}{|m{3.2cm}<{\centering}|}{Gu \etal \cite{gu2018efficient}}
	& \multicolumn{1}{m{1.6cm}<{\centering}}{6.76, 33.59}
	& \multicolumn{1}{m{1.6cm}<{\centering}|}{10.22, 26.97}
	& \multicolumn{1}{m{1.6cm}<{\centering}}{10.53, 33.22}
	& \multicolumn{1}{m{1.6cm}<{\centering}|}{10.81, 24.41}  
	& \multicolumn{1}{m{1.6cm}<{\centering}}{11.66, 24.31}
	& \multicolumn{1}{m{1.6cm}<{\centering}|}{12.64, 19.81}  \\
	
	\multicolumn{1}{|m{3.2cm}<{\centering}|}{Dabouei \etal \cite{dabouei2018fingerprint}}
	& \multicolumn{1}{m{1.6cm}<{\centering}}{6.70, 33.54}
	& \multicolumn{1}{m{1.6cm}<{\centering}|}{9.84, 27.16}
	& \multicolumn{1}{m{1.6cm}<{\centering}}{10.12, 33.82}
	& \multicolumn{1}{m{1.6cm}<{\centering}|}{10.53, 25.06}  
	& \multicolumn{1}{m{1.6cm}<{\centering}}{11.37, \underline{\textbf{24.53}}}
	& \multicolumn{1}{m{1.6cm}<{\centering}|}{12.35, 19.92}  \\
	
	\hline
	\multicolumn{1}{|m{3.2cm}<{\centering}|}{Proposed Method}
	& \multicolumn{1}{m{1.6cm}<{\centering}}{\textbf{6.44}, \textbf{33.76}}
	& \multicolumn{1}{m{1.6cm}<{\centering}|}{\textbf{9.46}, \textbf{27.57}}
	& \multicolumn{1}{m{1.6cm}<{\centering}}{\textbf{9.88}, \textbf{34.21}}
	& \multicolumn{1}{m{1.6cm}<{\centering}|}{\textbf{10.10}, \textbf{25.46}}  
	& \multicolumn{1}{m{1.6cm}<{\centering}}{\textbf{11.21}, 24.43}
	& \multicolumn{1}{m{1.6cm}<{\centering}|}{\textbf{12.18}, \textbf{20.06}}  \\
	
	\multicolumn{1}{|m{3.2cm}<{\centering}|}{Proposed Method +O}
	& \multicolumn{1}{m{1.6cm}<{\centering}}{\underline{\textbf{6.34}}, \underline{\textbf{33.77}}}
	& \multicolumn{1}{m{1.6cm}<{\centering}|}{\underline{\textbf{9.10}}, \underline{\textbf{27.82}}}
	& \multicolumn{1}{m{1.6cm}<{\centering}}{\underline{\textbf{9.60}}, \underline{\textbf{34.43}}}
	& \multicolumn{1}{m{1.6cm}<{\centering}|}{\underline{\textbf{9.72}}, \underline{\textbf{25.76}}}  
	& \multicolumn{1}{m{1.6cm}<{\centering}}{\underline{\textbf{11.17}}, \textbf{24.49}}
	& \multicolumn{1}{m{1.6cm}<{\centering}|}{\underline{\textbf{12.04}}, \underline{\textbf{20.28}}}  \\
	\hline
	
	\multicolumn{7}{l}{\emph{Note}: Elements of each column are marked according to their order as \underline{$\textbf{1}^\textbf{st}$} / $\textbf{2}^\textbf{nd}$ / others.}
	
	\end{tabular}
	\renewcommand{\arraystretch}{1}
\end{table*}

\begin{table}[!t]
	\caption{Distortion Estimation Accuracy of the Proposed Network with Different Submodules.\label{tab:ablation}}
	\centering
	\renewcommand{\arraystretch}{1.5}
	\begin{tabular}{|c c c c|c|}
		\hline
		\multicolumn{1}{|m{1.2cm}<{\centering}}{Spatial Pyramid Module} 
		& \multicolumn{1}{m{1.2cm}<{\centering}}{Attention Block} 
		& \multicolumn{1}{m{1.2cm}<{\centering}}{Mask} 
		& \multicolumn{1}{m{1.2cm}<{\centering}|}{Orientation Feature Branch}  
		& \multicolumn{1}{m{1.2cm}<{\centering}|}{$\mathcal{L}_{{\rm dis}}^{\rm{reg}}$}	\\
		\hline
		-			&-			&-			&-			& 109.6\\ 
		$\surd$ 	&- 			&- 			&- 			& 80.7\\
		$\surd$ 	&$\surd$	&- 			&- 			& 77.1\\
		$\surd$ 	&$\surd$	&$\surd$	&- 			& 75.1\\ \hline
		$\surd$ 	&$\surd$	&$\surd$	&$\surd$	& 69.6\\
		\hline
	\end{tabular}
	\renewcommand{\arraystretch}{1}
\end{table}

\begin{table}[!t]
	\caption{Distortion Estimation Accuracy of the Proposed Network with Different Preprocessing(enhancement, binarization, or thinning) and Training Stratigies(single-stage or two-stage).\label{tab:ablation_strategies}}
	\centering
	\renewcommand{\arraystretch}{1.5}
	\begin{tabular}{|c | c c c c|}
		\hline
		 \multicolumn{1}{|m{0.7cm}<{\centering}|}{} 
		& \multicolumn{1}{m{1.4cm}<{\centering}}{enhancement \& single-stage} 
		& \multicolumn{1}{m{1.4cm}<{\centering}}{binarization \& single-stage} 
		& \multicolumn{1}{m{1.4cm}<{\centering}}{thinning \quad\& single-stage} 
		& \multicolumn{1}{m{1.4cm}<{\centering}|}{thinning \quad\& \quad two-stage}\\
		\hline
		$\mathcal{L}_{{\rm dis}}^{\rm{reg}}$ & 	72.8 & 71.4	& 69.6	& 78.2 \\ 
		\hline
	\end{tabular}
	\renewcommand{\arraystretch}{1}
\end{table}

\subsection{Matching Performance}
To quantitatively evaluate the performance of rectification methods in a complete fingerprint recognition system, we further evaluated the matching performance using fingerprints after distortion rectification.
Four typical fingerprint recognition methods are applied to compute matching scores: 1) VeriFinger SDK 12.0 \cite{VeriFinger}, a widely used commercial software which is mainly based on minutiae features; 2) MCC \cite{cappelli2010minutia}, the state-of-the-art minutiae based matcher with published algorithm; 3) An in-house fixed length descriptor (IFLD); and 4) DeepPrint, a fast matching algorithm which extracts fixed-length descriptors for global texture features.
For convenience, we classify VeriFinger and MCC as minutiae based methods, IFLD and DeepPrint as fixed-length descriptor based methods, whose experimental results are shown respectively.
It should be noted that methods 3) and 4) extract a global descriptor for each fingerprint, which means that characteristics of some local areas may be ignored. 
Hence methods 1) and 2) are more suitable for evaluating the performance of distortion rectification, while methods 3) and 4) are only for reference.
Adopted matching strategies are described in Table \ref{tab:database}.

\begin{figure*}[!t]
	\centering
	\subfloat[]{\includegraphics[width=.33\linewidth]{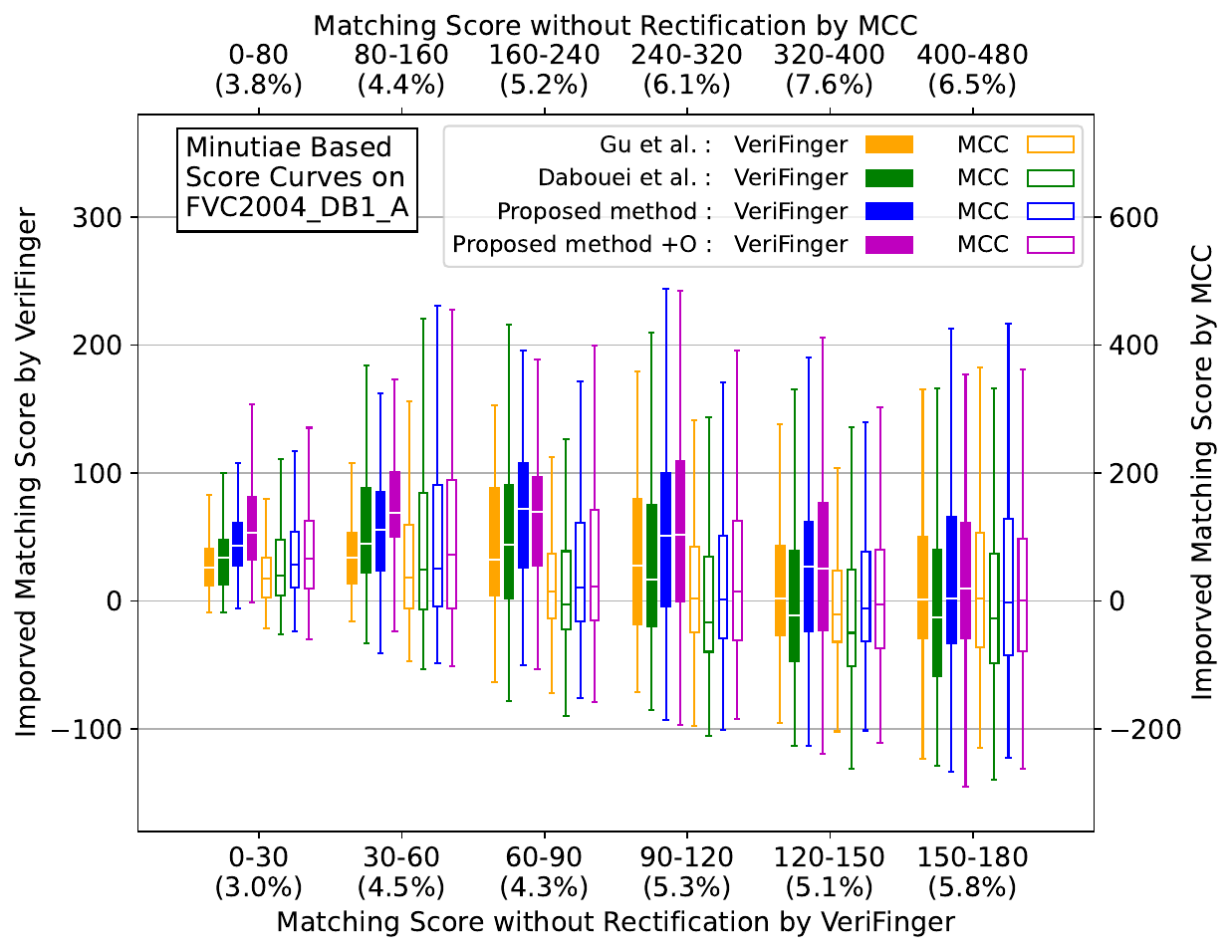}%
	}
	\hfil
	\subfloat[]{\includegraphics[width=.33\linewidth]{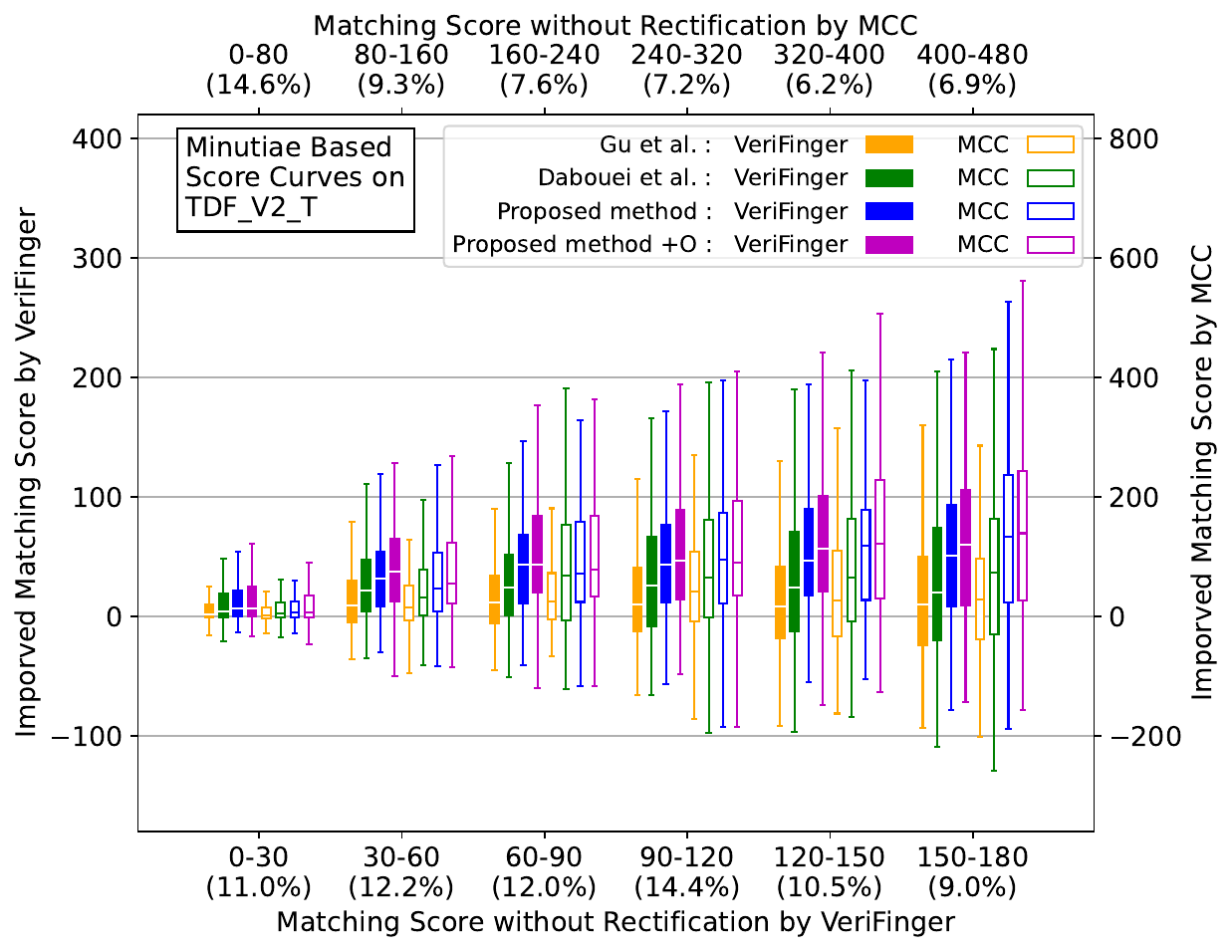}%
	}
	\hfil
	\subfloat[]{\includegraphics[width=.33\linewidth]{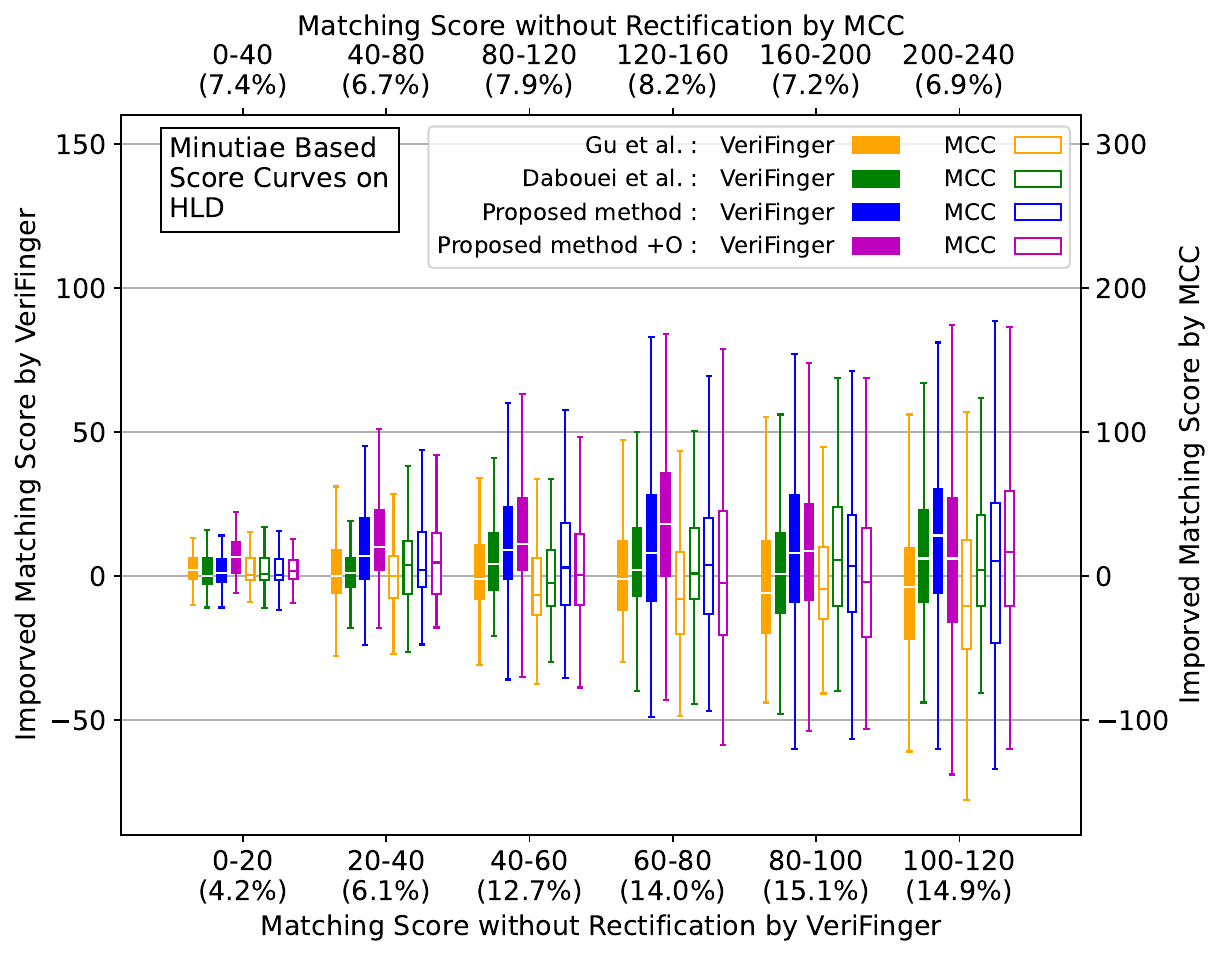}%
	}

	\subfloat[]{\includegraphics[width=.33\linewidth]{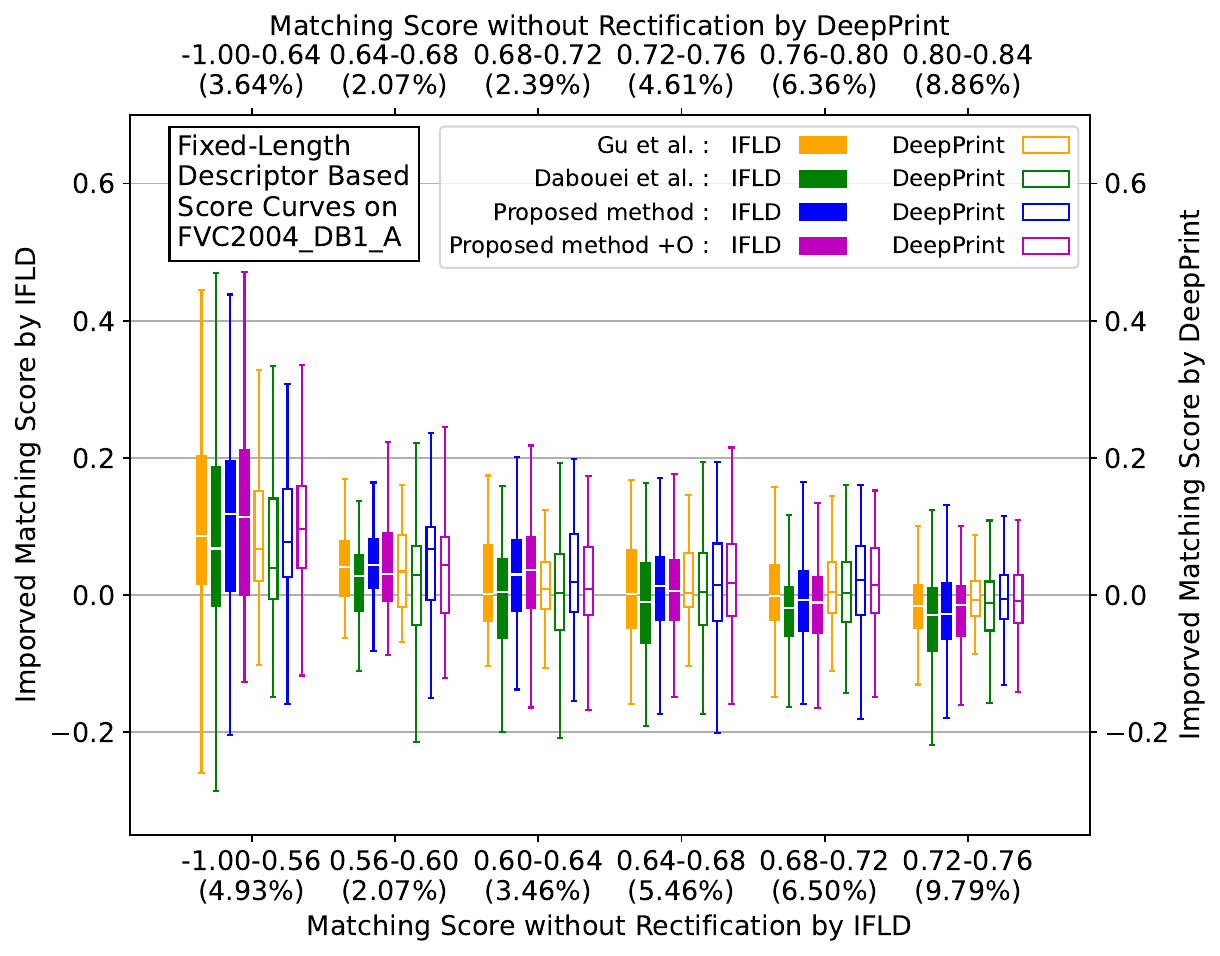}%
	}
	\hfil
	\subfloat[]{\includegraphics[width=.33\linewidth]{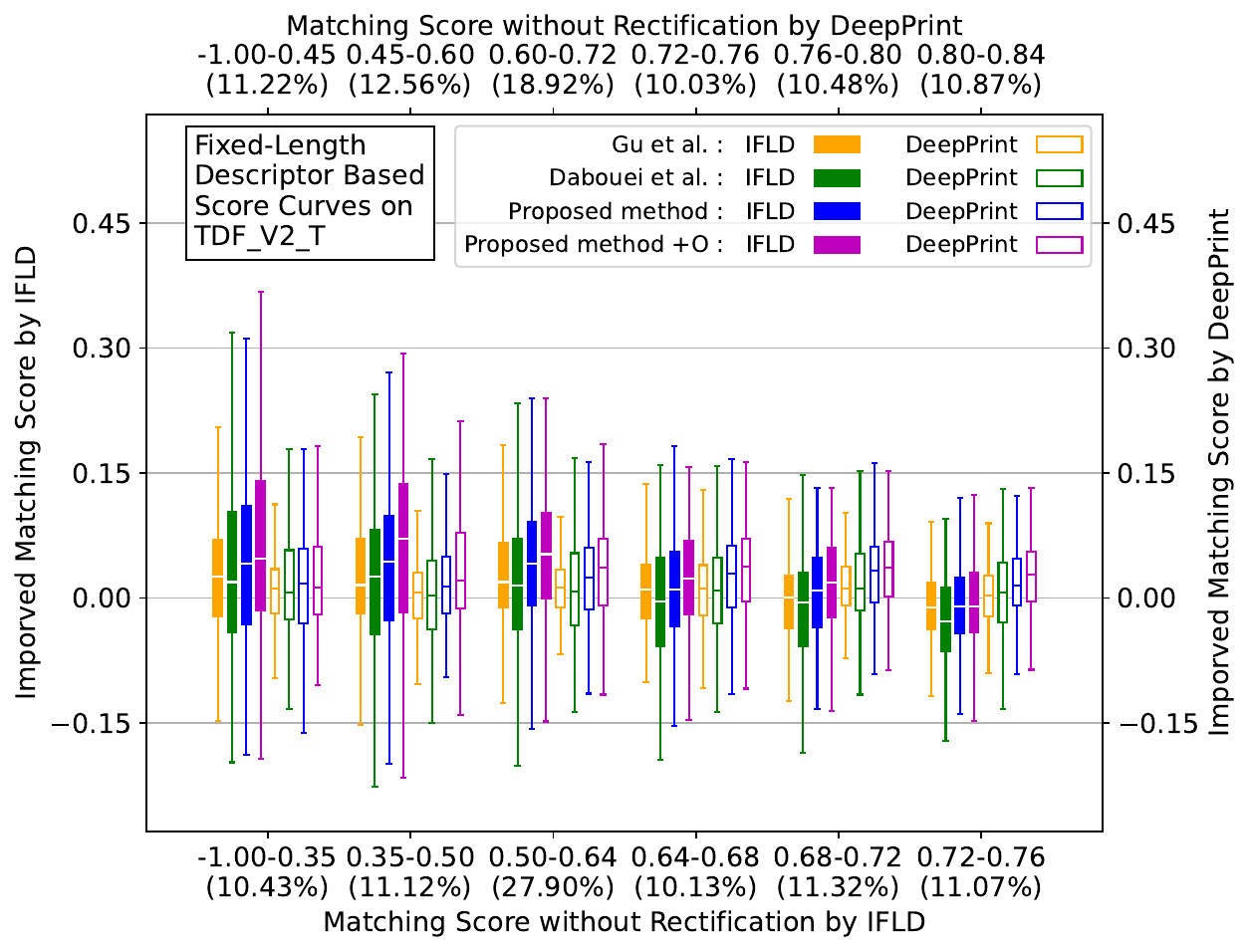}%
	}
	\hfil
	\subfloat[]{\includegraphics[width=.33\linewidth]{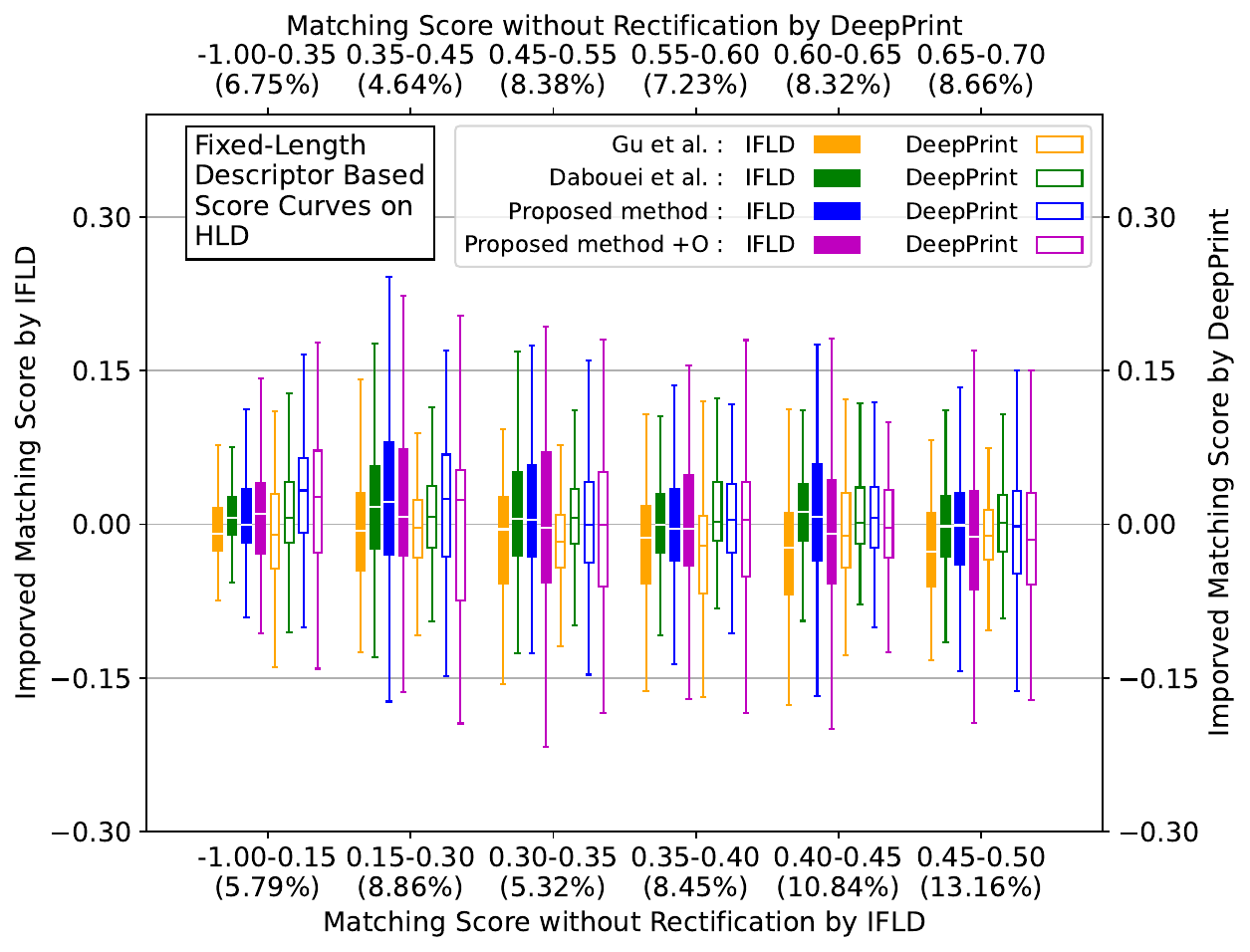}%
	}
	\caption{Boxplot of matching score improvements after rectification. The abscissa indicates the original matching score (without rectification) segment and its corresponding quantity ratio. The ordinate represents the distribution of improved scores after implementing different rectification algorithms.}
	\label{fig:improvement}
\end{figure*}

We first compare the matching score improvement of different rectification methods on FVC2004\_DB1\_A, TDF\_V2\_T and HLD. 
Only the pairs with relatively low original matching score (without rectification) are shown in Fig. \ref{fig:improvement}, since matching scores of the rest samples are sufficiently high for genuine match. 
The contrast in improved score distribution is more pronounced in minutiae based methods, which are more sensitive to distortions. 
From the distribution we can see that the proposed method performs better on most segments. PCA based methods do not work well in cases where the original matching scores are high (meaning that the fingerprint distortion is slight), because it is a global distortion and may incorrectly rectify some local areas.

\begin{table*}[!t]
	\caption{Identification Accuracy of Different Rectification Methods on TDF\_V2\_T. \label{tab:rank_TDFV2}}
	\centering
	\renewcommand{\arraystretch}{1.5}
	\begin{tabular}{|m{3.2cm}<{\centering}|m{1.2cm}<{\centering}|m{1.2cm}<{\centering}|m{1.2cm}<{\centering}|m{1.2cm}<{\centering}|m{1.2cm}<{\centering}|m{1.2cm}<{\centering}|m{1.2cm}<{\centering}|m{1.2cm}<{\centering}|}
		\hline
		\multirow{3}{*}{\textbf{Methods}} & \multicolumn{8}{c|}{\textbf{Rank 1 (\%) / Rank 5 (\%)}}\\
		\cline { 2 - 9 } 
		& \multicolumn{2}{m{2.4cm}<{\centering}|}{\textbf{VeriFinger\quad\;\;\,}}
		& \multicolumn{2}{m{2.4cm}<{\centering}|}{\textbf{MCC\quad\;\;\,}}  
		& \multicolumn{2}{m{2.4cm}<{\centering}|}{\textbf{IFLD\quad\;\;\,}}  
		& \multicolumn{2}{m{2.4cm}<{\centering}|}{\textbf{DeepPrint\quad\;\;\,}}\\
		\cline { 2 - 9 }
		& \multicolumn{1}{m{1.2cm}<{\centering}}{\textbf{full}}
		& \multicolumn{1}{m{1.2cm}<{\centering}|}{\textbf{subset}}
		& \multicolumn{1}{m{1.2cm}<{\centering}}{\textbf{full}}
		& \multicolumn{1}{m{1.2cm}<{\centering}|}{\textbf{subset}}
		& \multicolumn{1}{m{1.2cm}<{\centering}}{\textbf{full}}
		& \multicolumn{1}{m{1.2cm}<{\centering}|}{\textbf{subset}}
		& \multicolumn{1}{m{1.2cm}<{\centering}}{\textbf{full}}
		& \multicolumn{1}{m{1.2cm}<{\centering}|}{\textbf{subset}}  \\
		\hline
		\multicolumn{1}{|m{3.2cm}<{\centering}|}{Without Rectification}
		& \multicolumn{1}{m{1.2cm}<{\centering}}{89.3\,/\,91.3}
		& \multicolumn{1}{m{1.2cm}<{\centering}|}{77.6\,/\,82.5}
		& \multicolumn{1}{m{1.2cm}<{\centering}}{80.6\,/\,84.9}
		& \multicolumn{1}{m{1.2cm}<{\centering}|}{61.7\,/\,71.7}
		& \multicolumn{1}{m{1.2cm}<{\centering}}{89.7\,/\,93.4}
		& \multicolumn{1}{m{1.2cm}<{\centering}|}{76.4\,/\,85.0}
		& \multicolumn{1}{m{1.2cm}<{\centering}}{73.8\,/\,84.0}
		& \multicolumn{1}{m{1.2cm}<{\centering}|}{51.0\,/\,68.2}  \\
		
		\multicolumn{1}{|m{3.2cm}<{\centering}|}{Gu \etal \cite{gu2018efficient}}
		& \multicolumn{1}{m{1.2cm}<{\centering}}{90.7\,/\,92.6}
		& \multicolumn{1}{m{1.2cm}<{\centering}|}{81.3\,/\,85.7}
		& \multicolumn{1}{m{1.2cm}<{\centering}}{82.1\,/\,85.7}
		& \multicolumn{1}{m{1.2cm}<{\centering}|}{65.3\,/\,73.1}
		& \multicolumn{1}{m{1.2cm}<{\centering}}{90.3\,/\,94.3}
		& \multicolumn{1}{m{1.2cm}<{\centering}|}{78.4\,/\,86.9}
		& \multicolumn{1}{m{1.2cm}<{\centering}}{75.0\,/\,84.7}
		& \multicolumn{1}{m{1.2cm}<{\centering}|}{54.3\,/\,71.9}  \\
		
		\multicolumn{1}{|m{3.2cm}<{\centering}|}{Dabouei \etal \cite{dabouei2018fingerprint}}
		& \multicolumn{1}{m{1.2cm}<{\centering}}{91.3\,/\,93.0}
		& \multicolumn{1}{m{1.2cm}<{\centering}|}{82.3\,/\,86.2}
		& \multicolumn{1}{m{1.2cm}<{\centering}}{82.5\,/\,86.3}
		& \multicolumn{1}{m{1.2cm}<{\centering}|}{67.2\,/\,74.7}
		& \multicolumn{1}{m{1.2cm}<{\centering}}{90.6\,/\,93.5}
		& \multicolumn{1}{m{1.2cm}<{\centering}|}{79.6\,/\,86.2}
		& \multicolumn{1}{m{1.2cm}<{\centering}}{74.7\,/\,84.6}
		& \multicolumn{1}{m{1.2cm}<{\centering}|}{\textbf{55.6}\,/\,\textbf{73.5}}  \\
		\hline
		\multicolumn{1}{|m{3.2cm}<{\centering}|}{Proposed Method}
		& \multicolumn{1}{m{1.2cm}<{\centering}}{\textbf{91.9}\,/\,\textbf{94.0}}
		& \multicolumn{1}{m{1.2cm}<{\centering}|}{\textbf{83.2}\,/\,\textbf{88.6}}
		& \multicolumn{1}{m{1.2cm}<{\centering}}{\textbf{83.3}\,/\,\textbf{86.9}}
		& \multicolumn{1}{m{1.2cm}<{\centering}|}{\textbf{67.3}\,/\,\textbf{76.4}}
		& \multicolumn{1}{m{1.2cm}<{\centering}}{\textbf{91.4}\,/\,\textbf{94.5}}
		& \multicolumn{1}{m{1.2cm}<{\centering}|}{\textbf{81.0}\,/\,\textbf{88.1}}
		& \multicolumn{1}{m{1.2cm}<{\centering}}{\textbf{75.9}\,/\,\underline{\textbf{86.2}}}
		& \multicolumn{1}{m{1.2cm}<{\centering}|}{54.8\,/\,\underline{\textbf{74.1}}} \\
		
		\multicolumn{1}{|m{3.2cm}<{\centering}|}{Proposed Method +O}
		& \multicolumn{1}{m{1.2cm}<{\centering}}{\underline{\textbf{92.2}}\,/\,\underline{\textbf{94.3}}}
		& \multicolumn{1}{m{1.2cm}<{\centering}|}{\underline{\textbf{83.7}}\,/\,\underline{\textbf{89.3}}}
		& \multicolumn{1}{m{1.2cm}<{\centering}}{\underline{\textbf{83.9}}\,/\,\underline{\textbf{87.8}}}
		& \multicolumn{1}{m{1.2cm}<{\centering}|}{\underline{\textbf{69.4}}\,/\,\underline{\textbf{77.7}}}
		& \multicolumn{1}{m{1.2cm}<{\centering}}{\underline{\textbf{91.8}}\,/\,\underline{\textbf{94.8}}}
		& \multicolumn{1}{m{1.2cm}<{\centering}|}{\underline{\textbf{81.6}}\,/\,\underline{\textbf{89.1}}}
		& \multicolumn{1}{m{1.2cm}<{\centering}}{\underline{\textbf{77.8}}\,/\,\textbf{86.1}}
		& \multicolumn{1}{m{1.2cm}<{\centering}|}{\underline{\textbf{58.5}}\,/\,73.1}  \\
		\hline
		\multicolumn{7}{l}{\emph{Note}: Elements of each column are marked according to their order as \underline{$\textbf{1}^\textbf{st}$} / $\textbf{2}^\textbf{nd}$ / others.}
		
	\end{tabular}
	\renewcommand{\arraystretch}{1}
\end{table*}

\begin{table*}[!t]
	\caption{Identification Accuracy of Different Rectification Methods on HLD. \label{tab:rank_HLD}}
	\centering
	\renewcommand{\arraystretch}{1.5}
	\begin{tabular}{|m{3.2cm}<{\centering}|m{1.2cm}<{\centering}|m{1.2cm}<{\centering}|m{1.2cm}<{\centering}|m{1.2cm}<{\centering}|m{1.2cm}<{\centering}|m{1.2cm}<{\centering}|m{1.2cm}<{\centering}|m{1.2cm}<{\centering}|}
		\hline
		\multirow{3}{*}{\textbf{Methods}} & \multicolumn{8}{c|}{\textbf{Rank 1 (\%) / Rank 5 (\%)}}\\
		\cline { 2 - 9 } 
		& \multicolumn{2}{m{2.4cm}<{\centering}|}{\textbf{VeriFinger\quad\;\;\,}}
		& \multicolumn{2}{m{2.4cm}<{\centering}|}{\textbf{MCC\quad\;\;\,}}  
		& \multicolumn{2}{m{2.4cm}<{\centering}|}{\textbf{IFLD\quad\;\;\,}}  
		& \multicolumn{2}{m{2.4cm}<{\centering}|}{\textbf{DeepPrint\quad\;\;\,}}\\
		\cline { 2 - 9 }
		& \multicolumn{1}{m{1.2cm}<{\centering}}{\textbf{full}}
		& \multicolumn{1}{m{1.2cm}<{\centering}|}{\textbf{subset}}
		& \multicolumn{1}{m{1.2cm}<{\centering}}{\textbf{full}}
		& \multicolumn{1}{m{1.2cm}<{\centering}|}{\textbf{subset}}
		& \multicolumn{1}{m{1.2cm}<{\centering}}{\textbf{full}}
		& \multicolumn{1}{m{1.2cm}<{\centering}|}{\textbf{subset}}
		& \multicolumn{1}{m{1.2cm}<{\centering}}{\textbf{full}}
		& \multicolumn{1}{m{1.2cm}<{\centering}|}{\textbf{subset}}  \\
		\hline
		\multicolumn{1}{|m{3.2cm}<{\centering}|}{Without Rectification}
		& \multicolumn{1}{m{1.2cm}<{\centering}}{91.2\,/\,93.3}
		& \multicolumn{1}{m{1.2cm}<{\centering}|}{83.4\,/\,87.1}
		& \multicolumn{1}{m{1.2cm}<{\centering}}{89.6\,/\,92.8}
		& \multicolumn{1}{m{1.2cm}<{\centering}|}{80.7\,/\,86.4}
		& \multicolumn{1}{m{1.2cm}<{\centering}}{85.5\,/\,90.3}
		& \multicolumn{1}{m{1.2cm}<{\centering}|}{75.9\,/\,83.1}
		& \multicolumn{1}{m{1.2cm}<{\centering}}{70.6\,/\,83.6}
		& \multicolumn{1}{m{1.2cm}<{\centering}|}{58.7\,/\,74.8}  \\
		
		\multicolumn{1}{|m{3.2cm}<{\centering}|}{Gu \etal \cite{gu2018efficient}}
		& \multicolumn{1}{m{1.2cm}<{\centering}}{91.8\,/\,93.5}
		& \multicolumn{1}{m{1.2cm}<{\centering}|}{84.6\,/\,87.6}
		& \multicolumn{1}{m{1.2cm}<{\centering}}{\textbf{90.6}\,/\,93.4}
		& \multicolumn{1}{m{1.2cm}<{\centering}|}{\textbf{82.7}\,/\,87.3}
		& \multicolumn{1}{m{1.2cm}<{\centering}}{86.2\,/\,90.6}
		& \multicolumn{1}{m{1.2cm}<{\centering}|}{77.3\,/\,83.7}
		& \multicolumn{1}{m{1.2cm}<{\centering}}{71.6\,/\,83.0}
		& \multicolumn{1}{m{1.2cm}<{\centering}|}{60.3\,/\,74.2}  \\
		
		\multicolumn{1}{|m{3.2cm}<{\centering}|}{Dabouei \etal \cite{dabouei2018fingerprint}}
		& \multicolumn{1}{m{1.2cm}<{\centering}}{92.1\,/\,93.7}
		& \multicolumn{1}{m{1.2cm}<{\centering}|}{85.1\,/\,87.8}
		& \multicolumn{1}{m{1.2cm}<{\centering}}{90.3\,/\,93.4}
		& \multicolumn{1}{m{1.2cm}<{\centering}|}{81.9\,/\,87.3}
		& \multicolumn{1}{m{1.2cm}<{\centering}}{86.6\,/\,90.9}
		& \multicolumn{1}{m{1.2cm}<{\centering}|}{78.4\,/\,84.2}
		& \multicolumn{1}{m{1.2cm}<{\centering}}{71.7\,/\,83.8}
		& \multicolumn{1}{m{1.2cm}<{\centering}|}{60.0\,/\,\textbf{75.0}}  \\
		\hline
		\multicolumn{1}{|m{3.2cm}<{\centering}|}{Proposed Method}
		& \multicolumn{1}{m{1.2cm}<{\centering}}{\textbf{92.3}\,/\,\textbf{94.4}}
		& \multicolumn{1}{m{1.2cm}<{\centering}|}{\textbf{85.5}\,/\,\textbf{89.3}}
		& \multicolumn{1}{m{1.2cm}<{\centering}}{\textbf{90.6}\,/\,\textbf{93.5}}
		& \multicolumn{1}{m{1.2cm}<{\centering}|}{82.4\,/\,\textbf{87.7}}
		& \multicolumn{1}{m{1.2cm}<{\centering}}{\underline{\textbf{87.4}}\,/\,\underline{\textbf{91.4}}}
		& \multicolumn{1}{m{1.2cm}<{\centering}|}{\underline{\textbf{79.5}}\,/\,\underline{\textbf{85.1}}}
		& \multicolumn{1}{m{1.2cm}<{\centering}}{\textbf{72.3}\,/\,\textbf{84.2}}
		& \multicolumn{1}{m{1.2cm}<{\centering}|}{\textbf{61.2}\,/\,74.6} \\
		
		\multicolumn{1}{|m{3.2cm}<{\centering}|}{Proposed Method +O}
		& \multicolumn{1}{m{1.2cm}<{\centering}}{\underline{\textbf{93.2}}\,/\,\underline{\textbf{94.8}}}
		& \multicolumn{1}{m{1.2cm}<{\centering}|}{\underline{\textbf{87.2}}\,/\,\underline{\textbf{90.1}}}
		& \multicolumn{1}{m{1.2cm}<{\centering}}{\underline{\textbf{91.2}}\,/\,\underline{\textbf{93.9}}}
		& \multicolumn{1}{m{1.2cm}<{\centering}|}{\underline{\textbf{83.7}}\,/\,\underline{\textbf{88.5}}}
		& \multicolumn{1}{m{1.2cm}<{\centering}}{\textbf{87.0}\,/\,\underline{\textbf{91.4}}}
		& \multicolumn{1}{m{1.2cm}<{\centering}|}{\textbf{78.8}\,/\,\textbf{85.0}}
		& \multicolumn{1}{m{1.2cm}<{\centering}}{\underline{\textbf{72.6}}\,/\,\underline{\textbf{84.5}}}
		& \multicolumn{1}{m{1.2cm}<{\centering}|}{\underline{\textbf{61.7}}\,/\,\underline{\textbf{75.8}}}  \\
		\hline
		\multicolumn{7}{l}{\emph{Note}: Elements of each column are marked according to their order as \underline{$\textbf{1}^\textbf{st}$} / $\textbf{2}^\textbf{nd}$ / others.}
		
	\end{tabular}
	\renewcommand{\arraystretch}{1}
\end{table*}

Table \ref{tab:rank_TDFV2} and Table \ref{tab:rank_HLD} shows the Rank-1 and Rank-5 identification accuracies of four representative matchers on both full set and subset of TDF\_V2\_T and HLD. 
FVC2004\_DB1\_A is not used because of its limited number of fingers. 
It can be observed that our method outperforms other rectification methods on various matchers based on different features.

\begin{figure*}[!t]
	\centering
	\subfloat[]{\includegraphics[width=.33\linewidth]{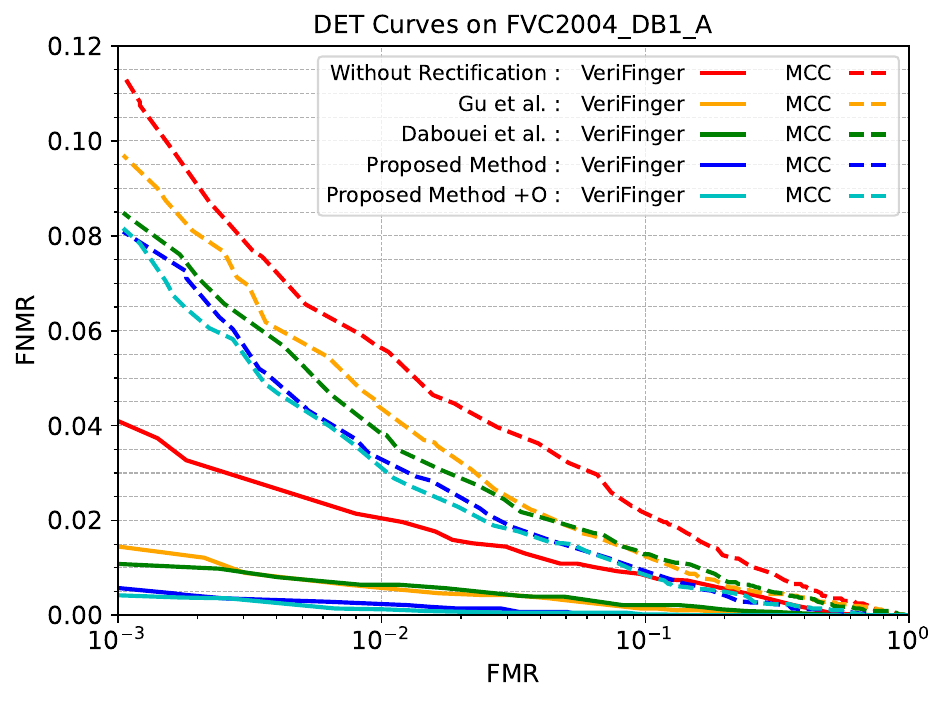}%
	}
	\hfil
	\subfloat[]{\includegraphics[width=.33\linewidth]{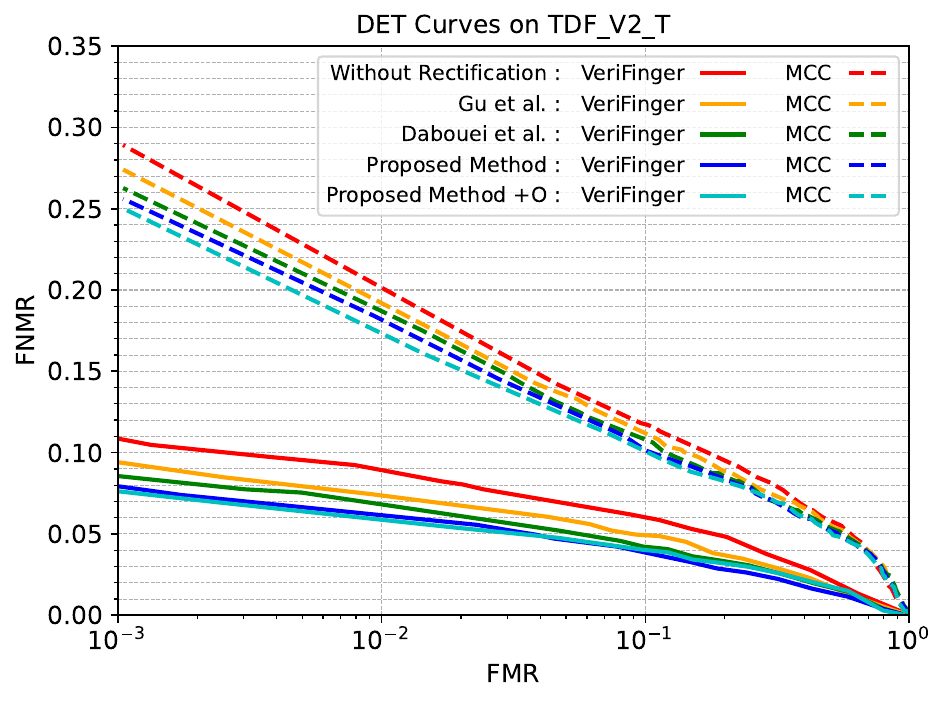}%
	}
	\hfil
	\subfloat[]{\includegraphics[width=.33\linewidth]{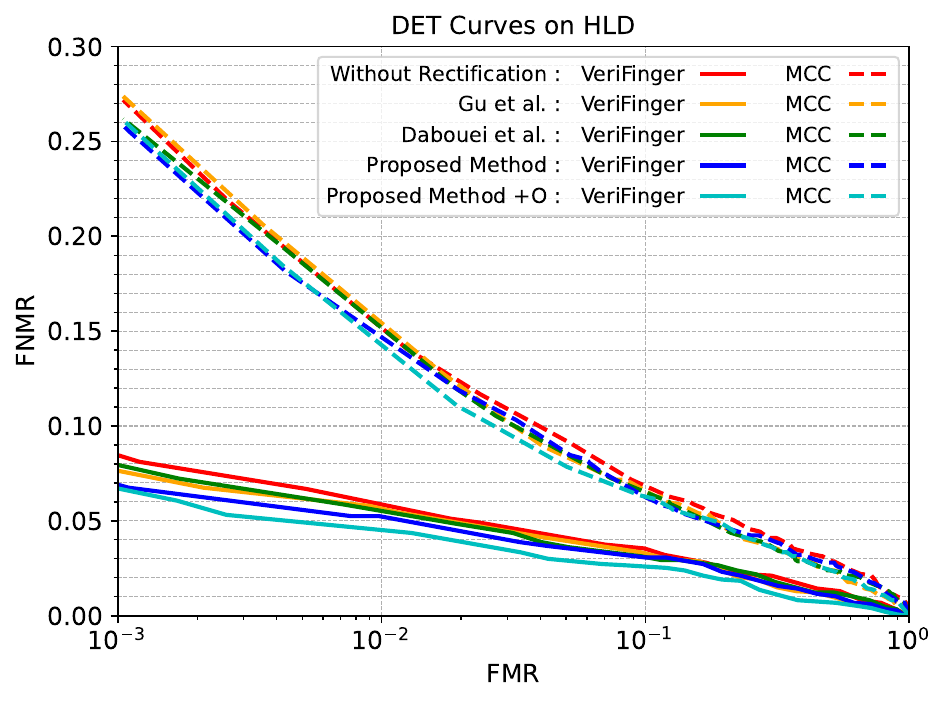}%
	}
	
	\subfloat[]{\includegraphics[width=.33\linewidth]{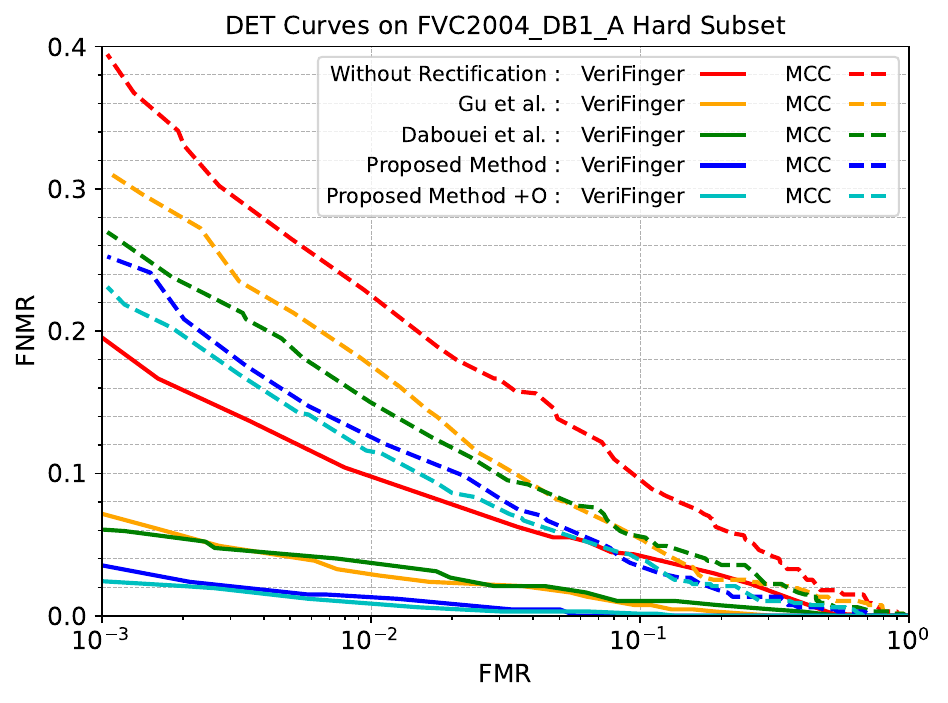}%
	}
	\hfil
	\subfloat[]{\includegraphics[width=.33\linewidth]{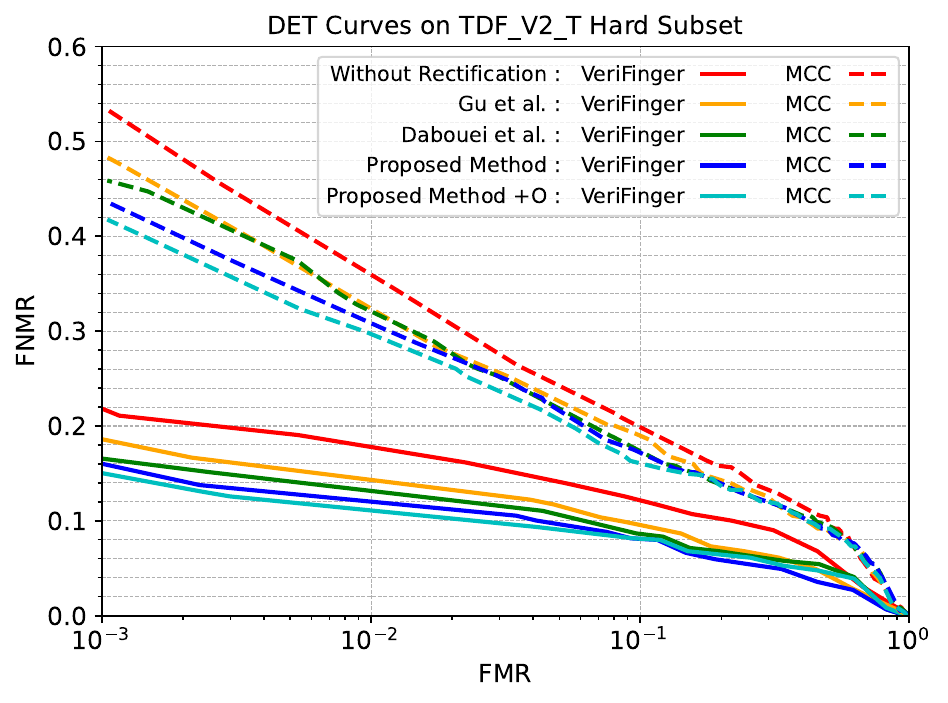}%
	}
	\hfil
	\subfloat[]{\includegraphics[width=.33\linewidth]{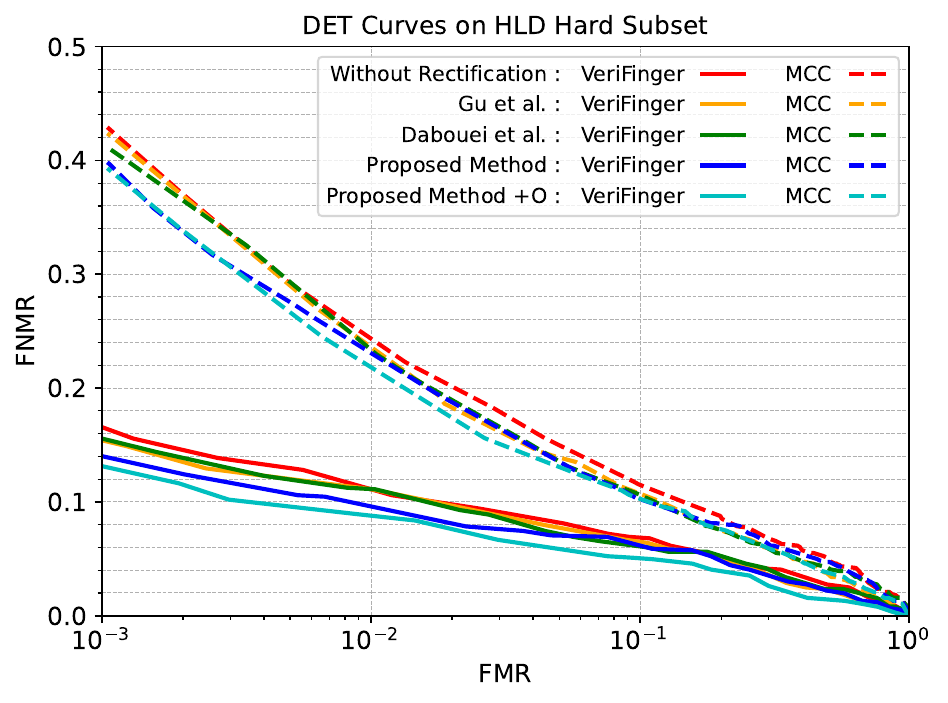}%
	}
	\caption{DET curves by minutiae based matchers with different rectification methods on three databases: FVC2004\_DB1\_A, TDF\_V2\_T, and HLD. Results of the full set and its corresponding hard subset are listed on the upper and lower sides respectively.}
	\label{fig:DET_minutiae}
\end{figure*}

\begin{figure*}[!t]
	\centering
	\subfloat[]{\includegraphics[width=.33\linewidth]{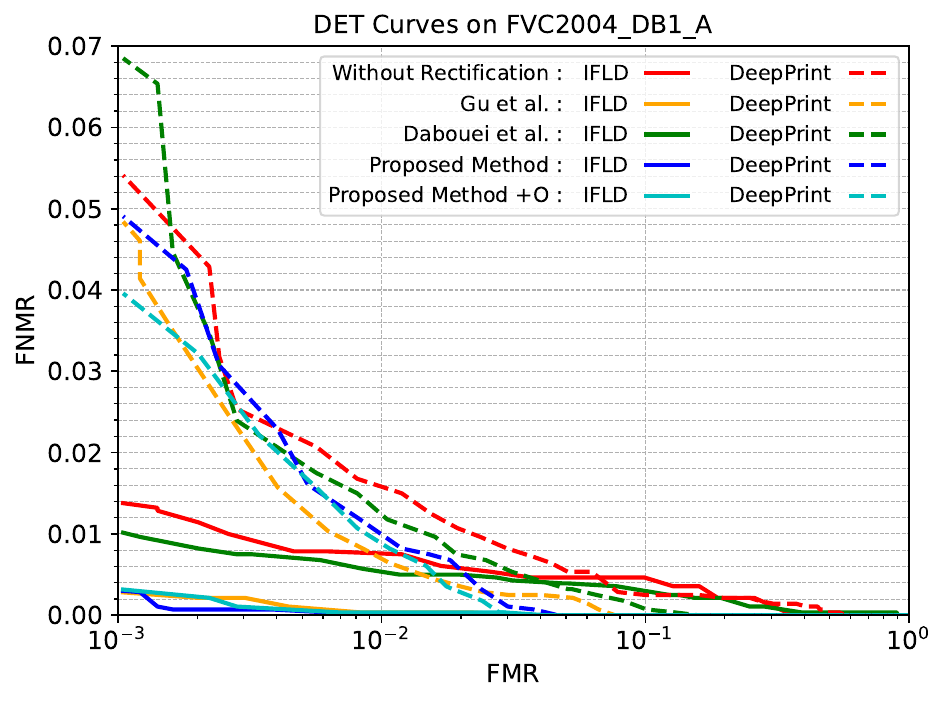}%
	}
	\hfil
	\subfloat[]{\includegraphics[width=.33\linewidth]{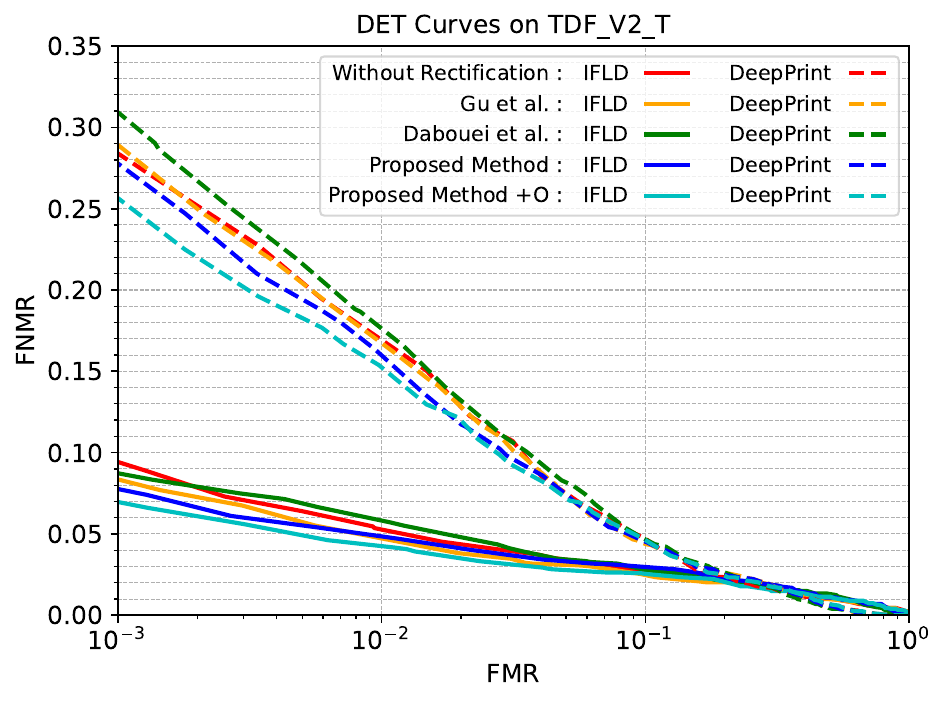}%
	}
	\hfil
	\subfloat[]{\includegraphics[width=.33\linewidth]{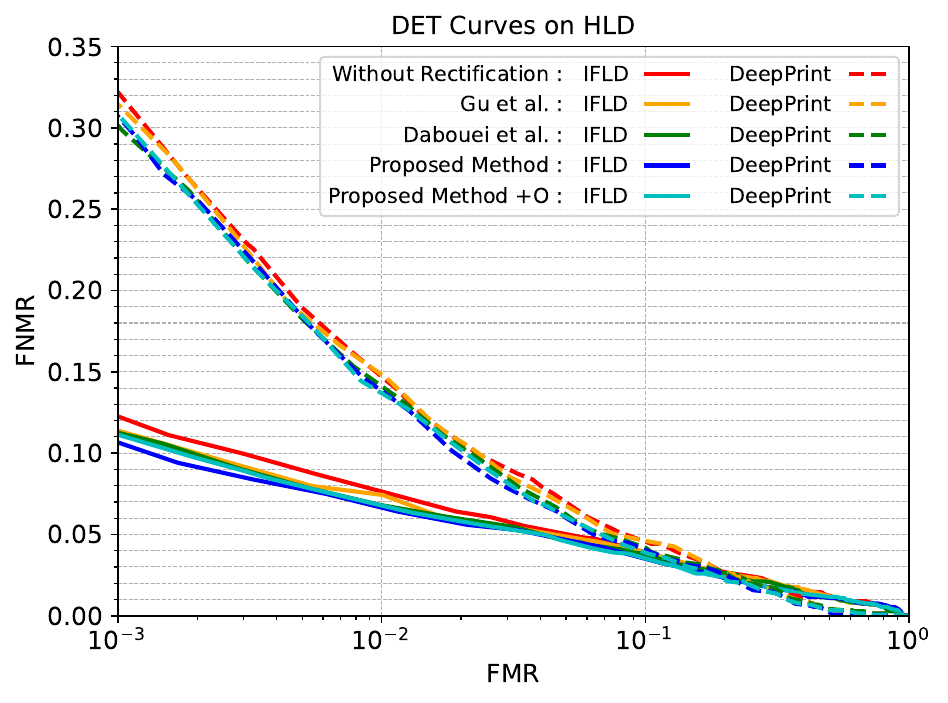}%
	}
	
	\subfloat[]{\includegraphics[width=.33\linewidth]{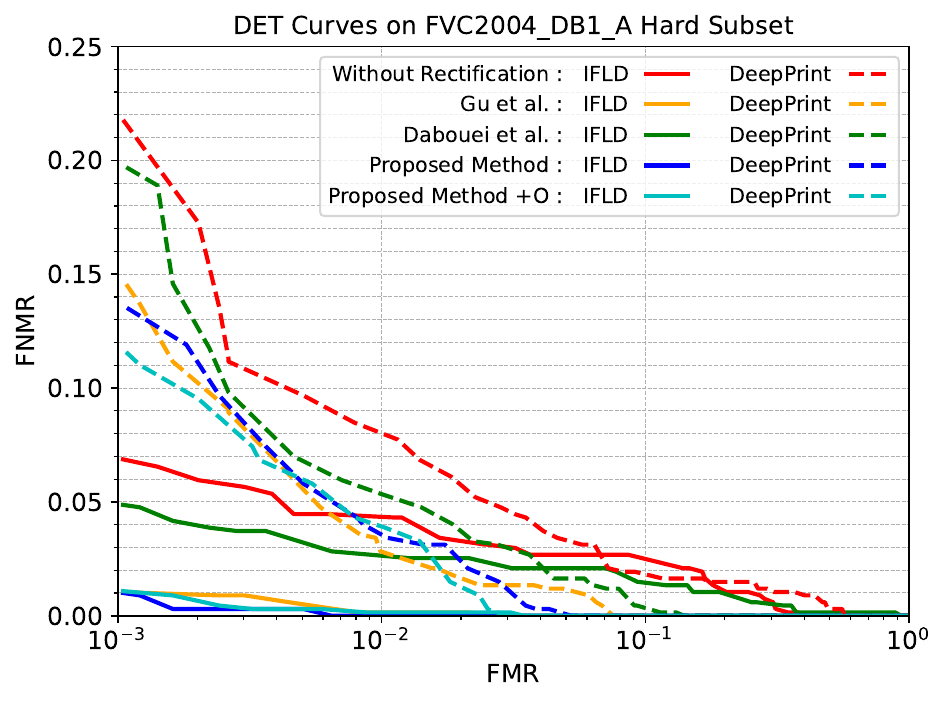}%
	}
	\hfil
	\subfloat[]{\includegraphics[width=.33\linewidth]{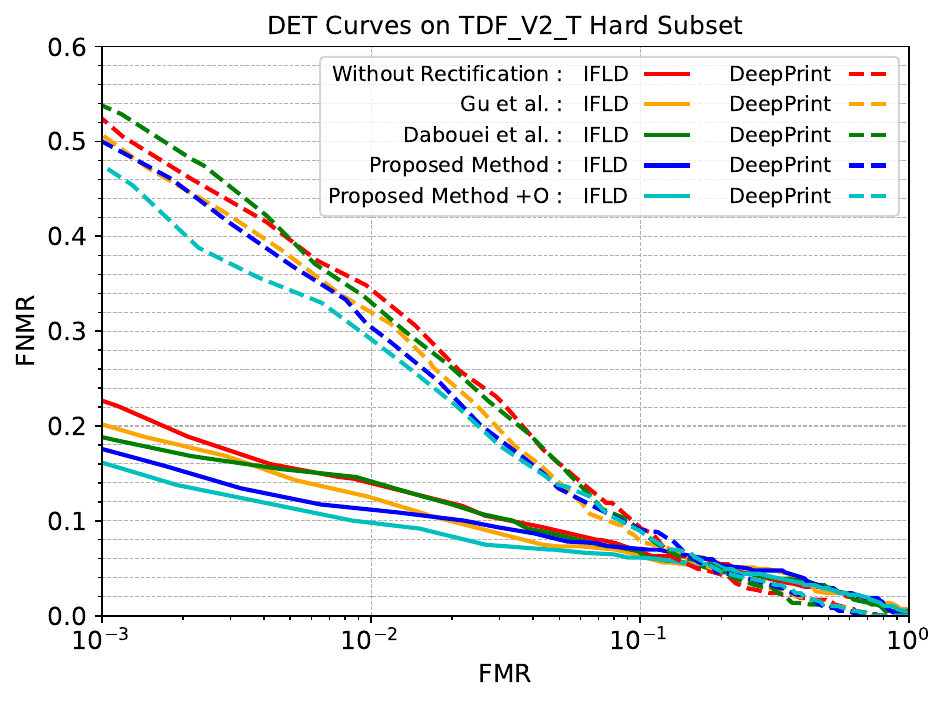}%
	}
	\hfil
	\subfloat[]{\includegraphics[width=.33\linewidth]{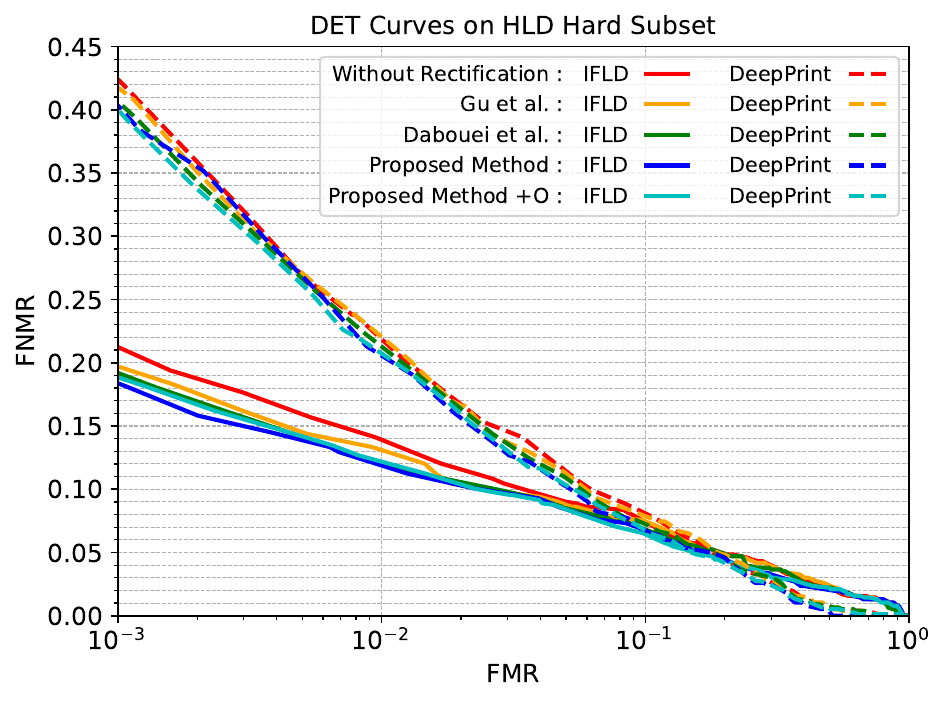}%
	}
	\caption{DET curves by fixed-length descriptor based matchers with different rectification methods on three databases: FVC2004\_DB1\_A, TDF\_V2\_T, and HLD. Results of the full set and its corresponding hard subset are listed on the upper and lower sides respectively.}
	\label{fig:DET_descriptor}
\end{figure*}

\begin{figure*}[!t]
	\centering
	\subfloat[]{\includegraphics[width=.4\linewidth]{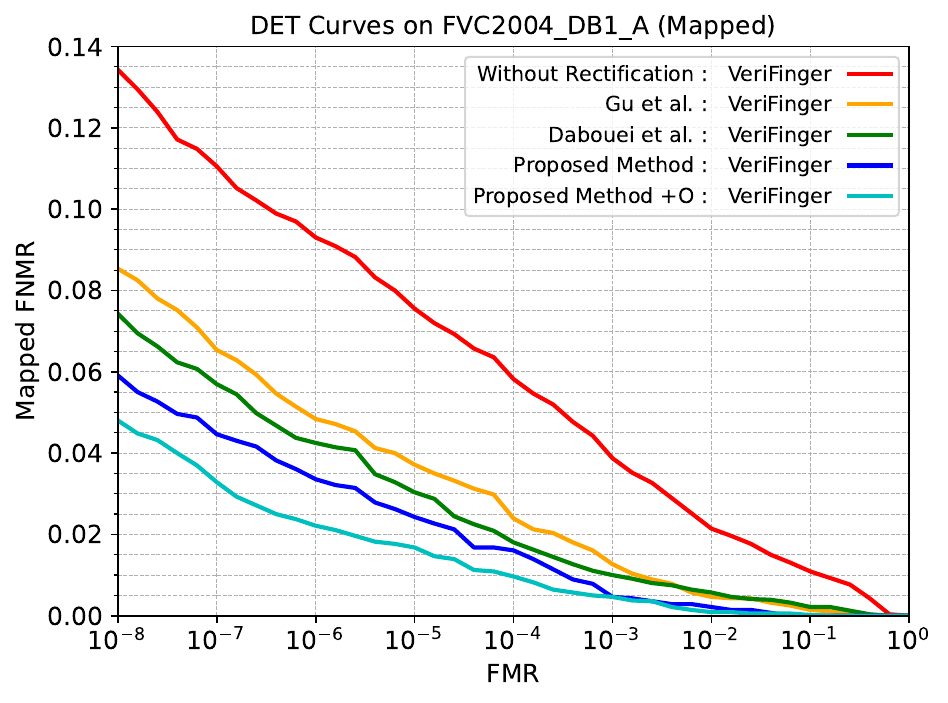}%
	}
	\hfil
	\subfloat[]{\includegraphics[width=.4\linewidth]{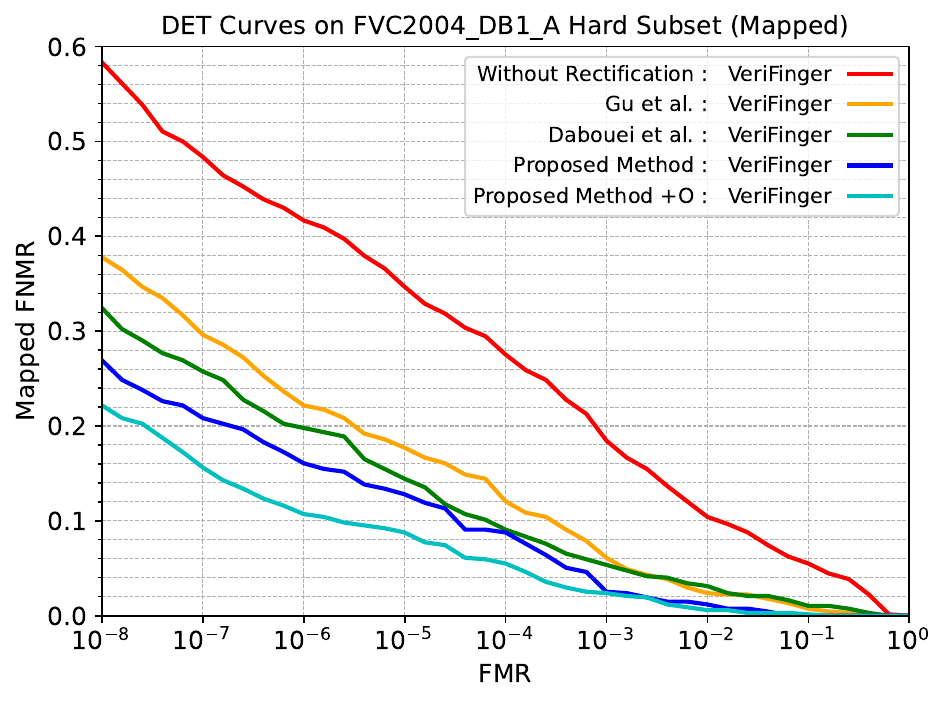}%
	}
	
	\subfloat[]{\includegraphics[width=.4\linewidth]{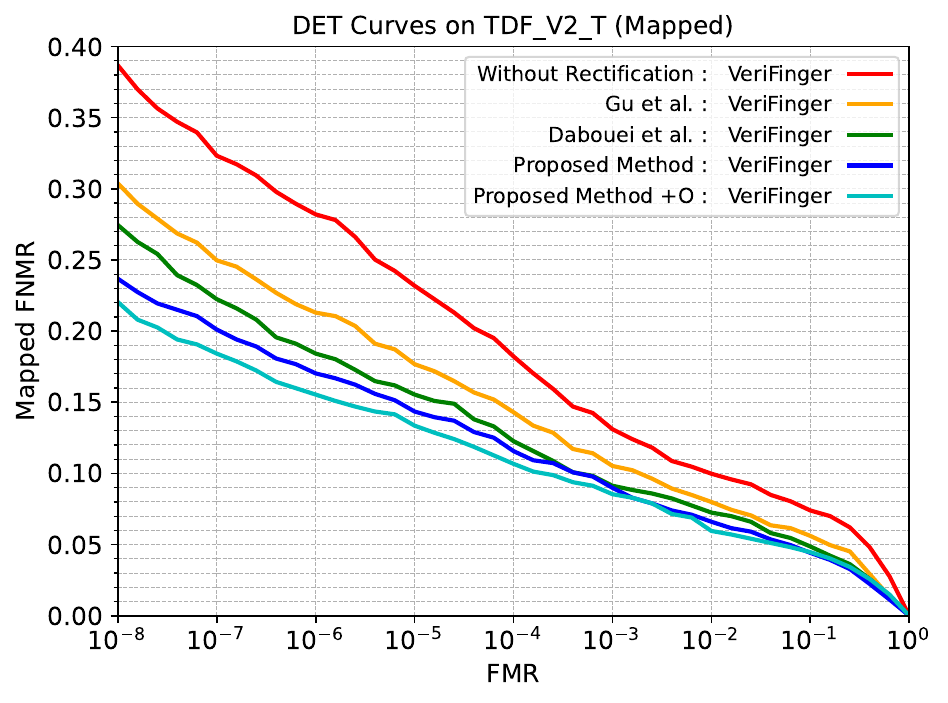}%
	}
	\hfil
	\subfloat[]{\includegraphics[width=.4\linewidth]{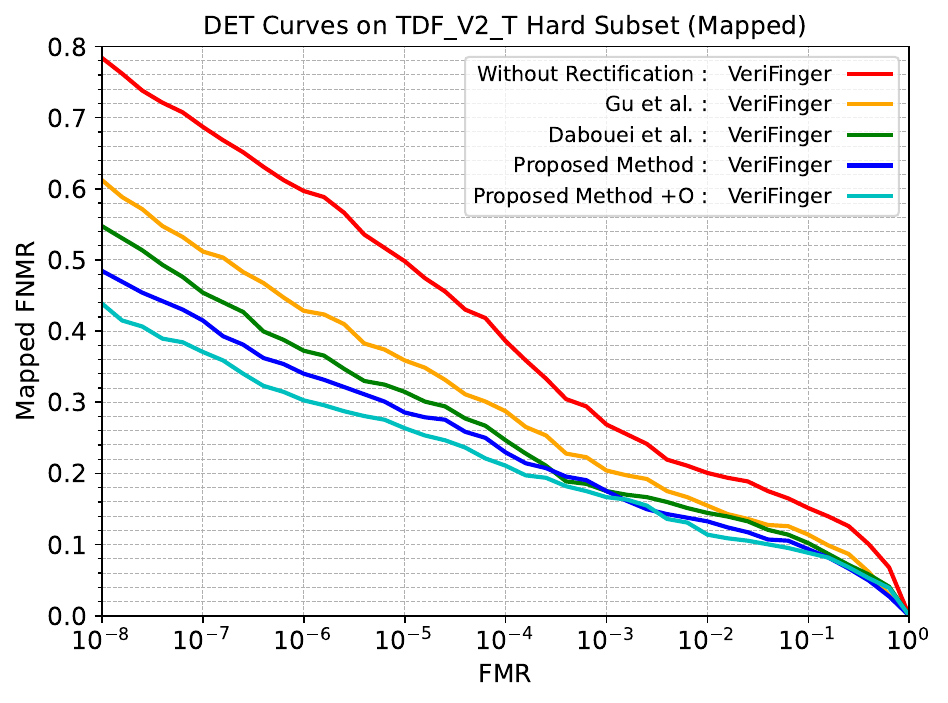}%
	}
	
	\subfloat[]{\includegraphics[width=.4\linewidth]{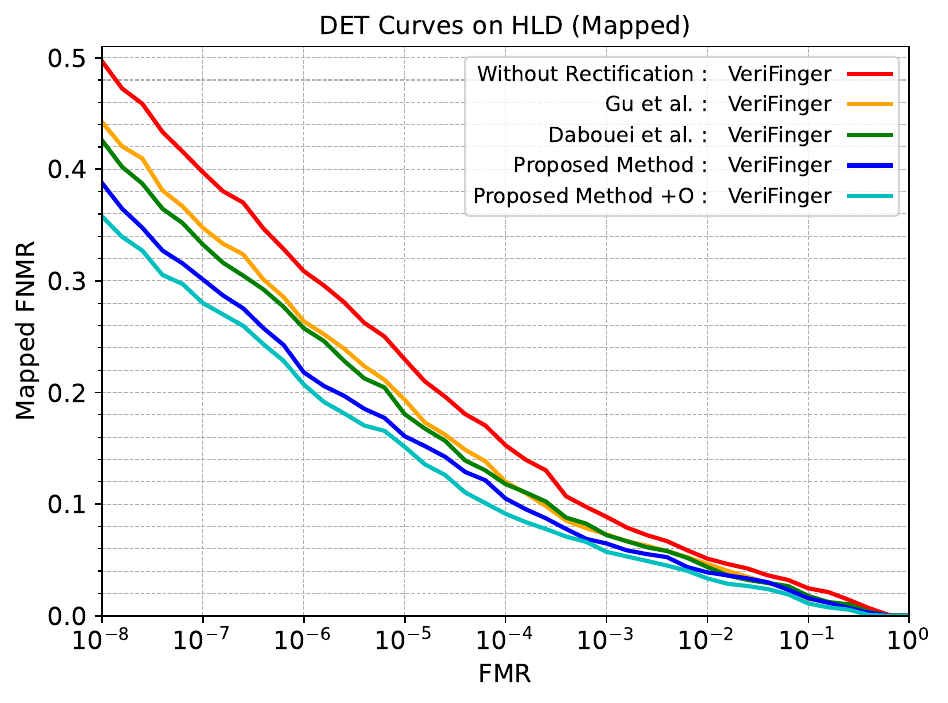}%
	}
	\hfil
	\subfloat[]{\includegraphics[width=.4\linewidth]{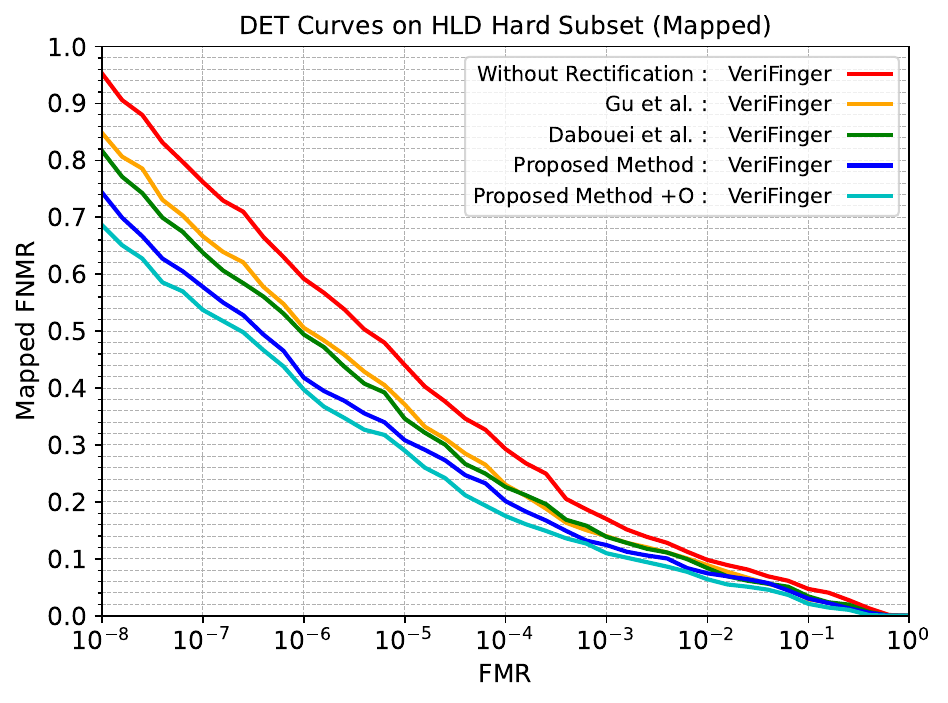}%
	}
	\caption{Mapped DET curves by VeriFinger matcher with different rectification methods on three databases: FVC2004\_DB1\_A, TDF\_V2\_T, and HLD. Results of the full set and its corresponding hard subset are listed on the left and right sides respectively.}
	\label{fig:DET_verifinger}
\end{figure*}

To further evaluate the rectification performance in recognition task, Detection Error Tradeoff (DET) curves are depicted on FVC2004\_DB1\_A,  TDF\_V2\_T, HLD and their hard subsets.
Fig. \ref{fig:DET_minutiae} and Fig. \ref{fig:DET_descriptor} display the plots by minutiae based and fixed-length descriptor based matchers respectively.
Due to limited data, the lowest FMR is only displayed to $10^{-3}$.
In minutiae based matching experiments shown in Fig. \ref{fig:DET_minutiae}, it can be seen that both of our proposed methods outperform previous PCA-based methods, while show little difference between each other in this small amount.
The fixed-length descriptor based curve analysis shown in Fig. \ref{fig:DET_descriptor} is similar to Fig. \ref{fig:DET_minutiae}.
Besides, the proposed method with orientation feature is significantly better than other methods in subfigures (a)(b)(d), which proves the complete network rectifies the texture structure of each fingerprint area more reasonably rather than just focusing on locations with a large distortion and rectifying stiffly.
Curves on subfigures (c)(f) are basically the same due to the complex and noisy background of images in HLD, while fixed-length descriptor based matchers are not specially trained for it.

In addition, we also obtain mapped DET curves using genuine matching scores of VeriFinger, which has been designed to map the false match rate (FMR), same as previous experiments \cite{si2015detection,gu2018efficient,dabouei2018fingerprint}.
This allows us to measure the false non-match rates (FNMR) at lower FMRs despite the limited number of imposter matches. 
The mapped DET curves are shown in Fig. \ref{fig:DET_verifinger}.
It can be observed from these curves that our method significantly exceeds other rectification methods on the matching performance at very low FMR region.
There is an additional improvement between our proposed complete and basic network, which also demonstrates that orientation features can indeed supplement some missing distortion information.

\begin{figure}[!t]
	\centering
	\includegraphics[width=.95\linewidth]{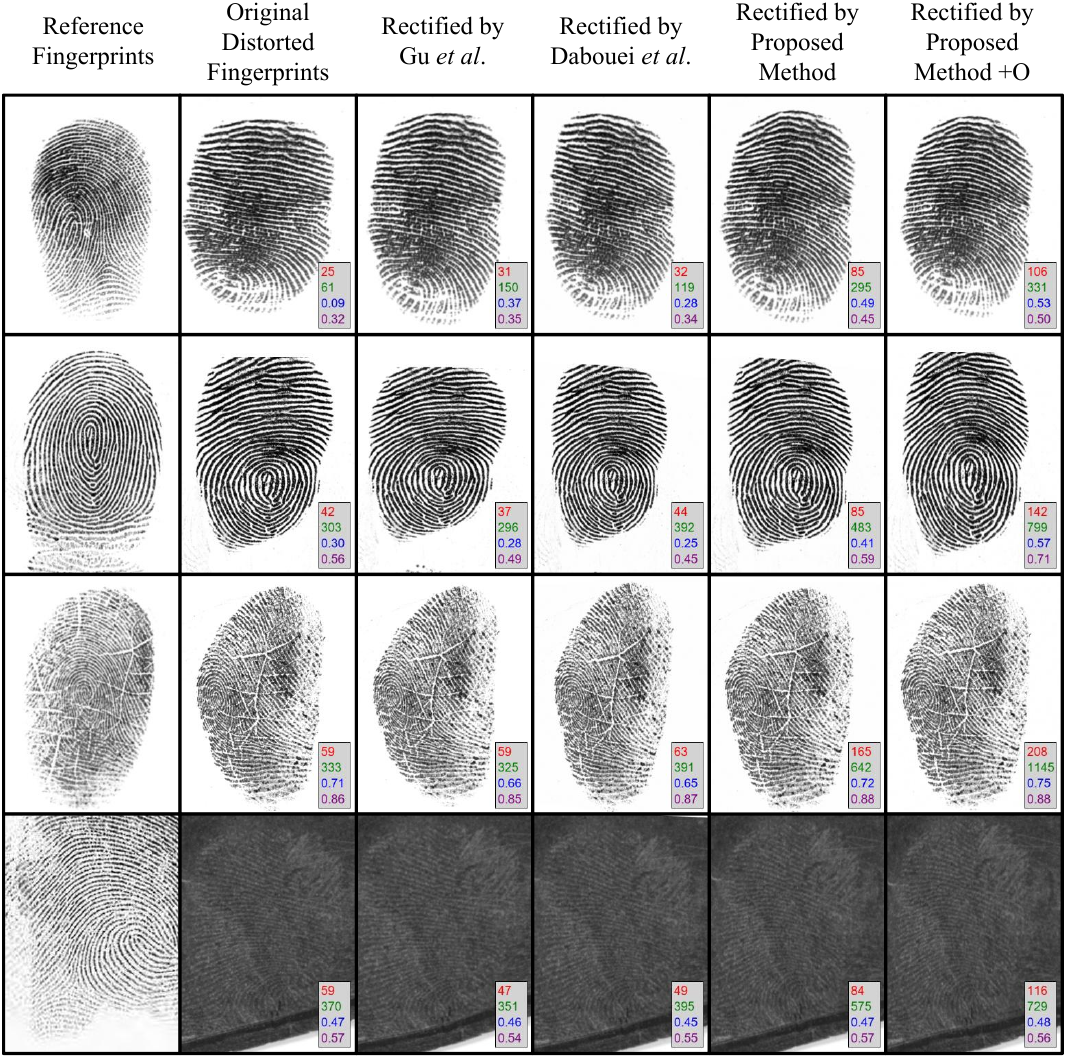}
	\caption{Examples of different rectification methods. Numbers in red$\backslash$green$\backslash$blue$\backslash$purple are matching scores between the normal fingerprint and the rectified results calculated by VeriFinger$\backslash$MCC$\backslash$IFLD$\backslash$DeepPrint.}
	\label{fig:examples_good}
\end{figure} 

\begin{figure}[!t]
	\centering
	\includegraphics[width=.95\linewidth]{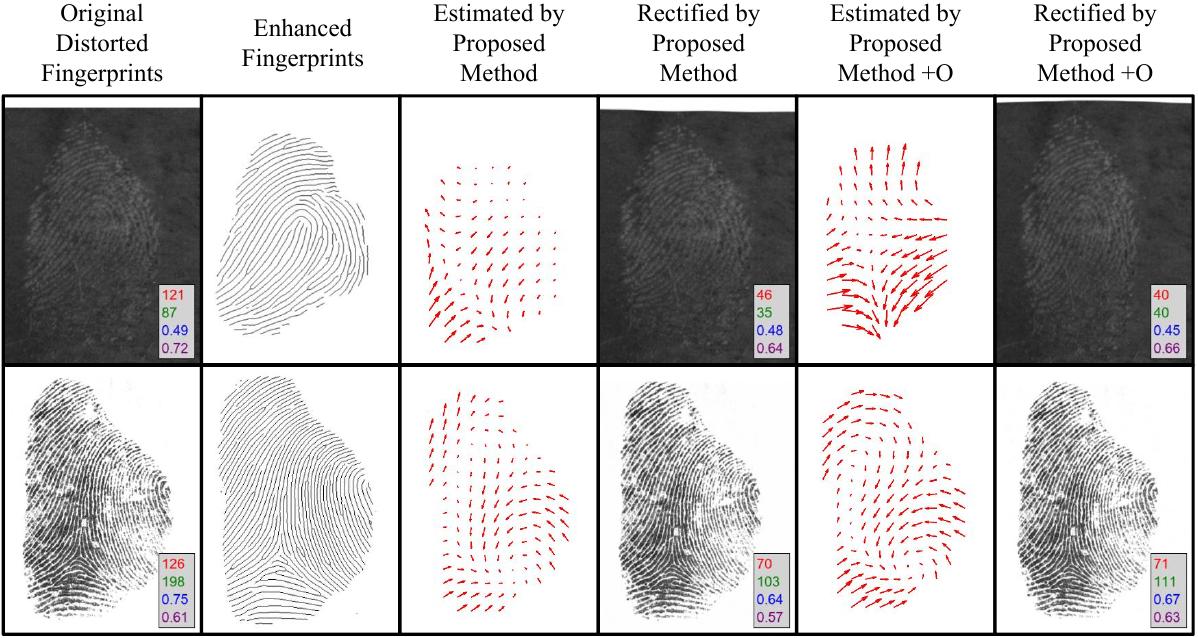}
	\caption{Two failure rectification examples by the proposed method. Numbers in red$\backslash$green$\backslash$blue$\backslash$purple are matching scores between the normal fingerprint and the rectified results calculated by VeriFinger$\backslash$MCC$\backslash$IFLD$\backslash$DeepPrint.}
	\label{fig:examples_bad}
\end{figure} 

Four examples are given in Fig. \ref{fig:examples_good} to compare the rectified results, which represent the following scenarios from top to bottom: (1) central distortion in up pose; (2) noncentral distortion in front pose; (3) noncentral distrotion in side pose; (4) complex distortion with background noise.
Score distributions of different matchers in corresponding low segments are listed in abscissas of Fig. \ref{fig:improvement}.
In these cases, existing methods expose the following problems:  (1) hard
to deal with fingerprints whose pose cannot be accurately normalized (line 1,2,3); (2) only significant distortion in one direction is concerned (marginal distortion is ignored in line 1); (3) cannot infer partial or complex distortion (line 3 and 4), while our method works well in these cases.

As shown in Fig. \ref{fig:examples_bad}, our rectification method still fails in some cases.
On the first line, a latent fingerprint is wrongly preprocessed (fake texture added on the bottom-right region), which leads to misjudgments during distortion estimation.
The second line shows an example with a very lateral angle, in which the area close to knuckle is considered distorted and wrongly compressed.

\begin{table}[h]
	\caption{Model Size and Efficiency of Different Rectification Methods for Processing a $512\times512$ Distorted Fingerprint in TDF\_V2\_T.\label{tab:cost}}
	\centering
	\renewcommand{\arraystretch}{1.5}
	\begin{tabular}{|l|c|c|}
		\hline
		Methods                      		& Params (M) & Time (s) \\
		\hline
		Gu \etal \cite{gu2018efficient}		& 112.5	& 3.85	\\
		Dabouei \etal \cite{dabouei2018fingerprint}	& 63.0	& 0.39	\\
		\hline
		Proposed		& 11.4	& 0.36	\\
		Proposed +O                               	&  45.0 &  0.41 \\
		\hline
	\end{tabular}
	\renewcommand{\arraystretch}{1}
\end{table}
\subsection{Efficiency Analysis}
Table \ref{tab:cost} shows the model size and efficiency of different rectification algorithms. Parameters of \cite{gu2018efficient} and \cite{dabouei2018fingerprint} contain the principal distortion patterns extracted by PCA. 
Especially, the parameters of Gu \etal\ method \cite{gu2018efficient} are related to the amount of training data, where $5000$ samples are used in our experiments.
Time in this table is the average time from inputting a $512\times512$ distorted fingerprint in TDF\_V2\_T to outputting its rectified result. 
It can be seen that our method with the basic network has the smallest number of parameters compared to previous methods, and at the same time has better rectification effect.
Moreover, when using the full network, the corresponding model size is acceptable while further improving performance.
There is little difference in time cost except for Gu \etal\ method, in which additional extraction of local ridge orientation and period is required.
All rectification algorithms are implemented in Python on a computer with a NVIDIA GeForce RTX 2080 Ti GPU and a 1.2 GHz CPU.

\section{Conclusion}\label{sec:conclusion}
In this paper, we propose a fingerprint distortion rectification algorithm where the dense distortion field of a single fingerprint is directly regressed. 
Self-reference relationship combining multiple features is constructed in the proposed network to finely estimate the detailed distortion patterns, instead of a sparse combination of principle components.
Moreover, the performance of our proposed method does not depend on pose normalization, which is unreliable for fingerprints with several distortion and non-frontal poses.
More distorted fingerprints with diverse poses and various distortion types are collected to make several more challenging and diverse distorted fingerprint databases.
Experiments show our proposed method outperforms state-of-the-art rectification algorithms.

The limitation of the proposed method is that input fingerprints are preprocessed in advance, which overcomes most of the effects caused by poor image quality but sometimes still fails with preprocessing errors.
In addition, it is more sensitive to local textures compared with PCA-based methods and may make mistakes on some rare and specific fingerprint patterns.
 
In the future we will focus on: (1) combining prior information with self-reference to estimate distortion field; (2) extending our method to more challenging fingerprints with incomplete or small areas; (3) considering 3D constraints instead of 2D deformations, which are more realistic.

{
\bibliographystyle{IEEEtran}
\bibliography{egbib}{}
}

\begin{IEEEbiography}[{\includegraphics[width=1in,height=1.25in,clip,keepaspectratio]{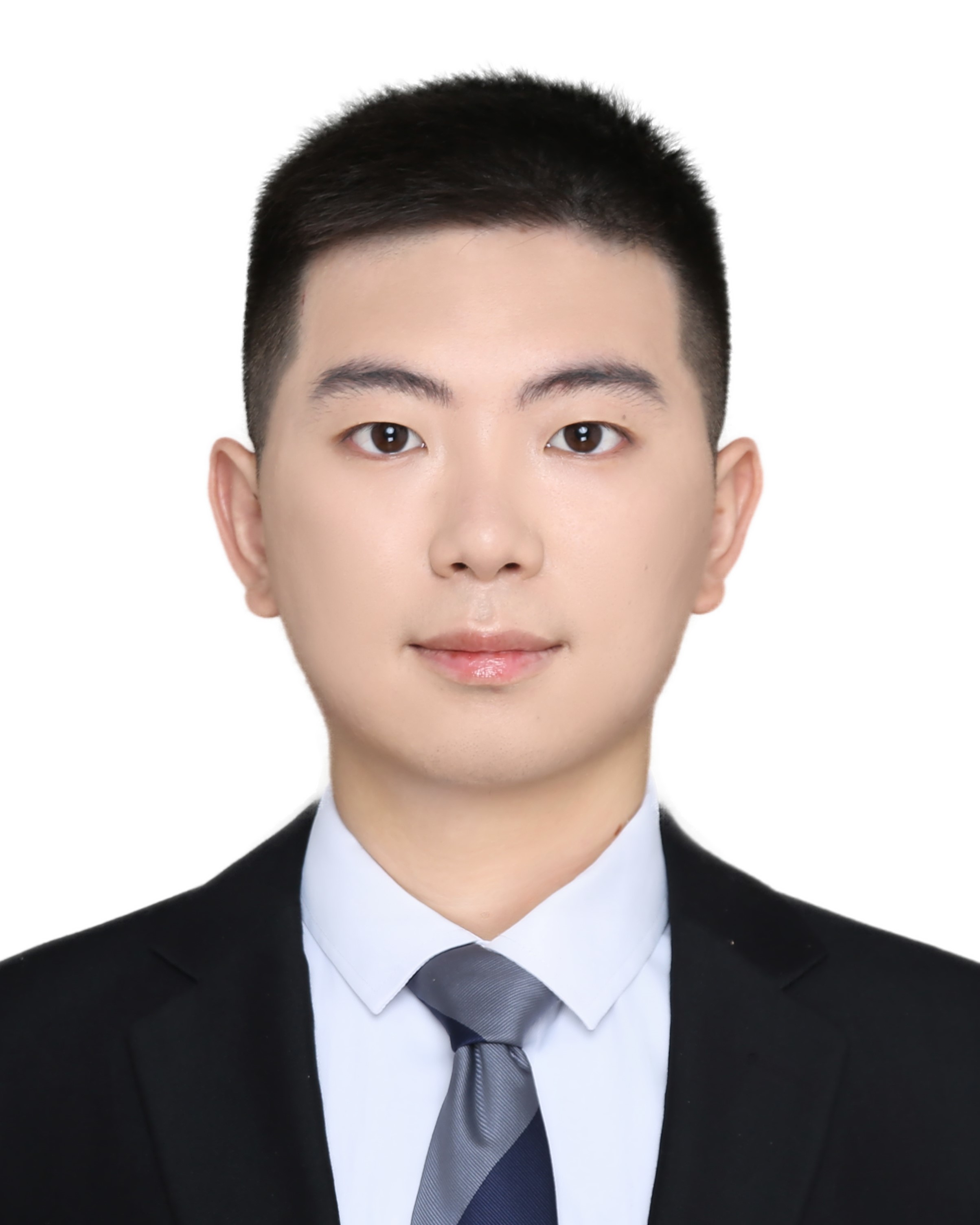}}]{Xiongjun Guan} received the B.S. degree from the Department of Automation, Tsinghua University, Beijing,
	China, in 2021, where he is currently pursuing the Ph.D. degree under the supervision of Prof.
	Jianjiang Feng with the Department of Automation. His research interests include fingerprint
	recognition, computer vision and pattern recognition.
\end{IEEEbiography}

\begin{IEEEbiography}[{\includegraphics[width=1in,height=1.25in,clip,keepaspectratio]{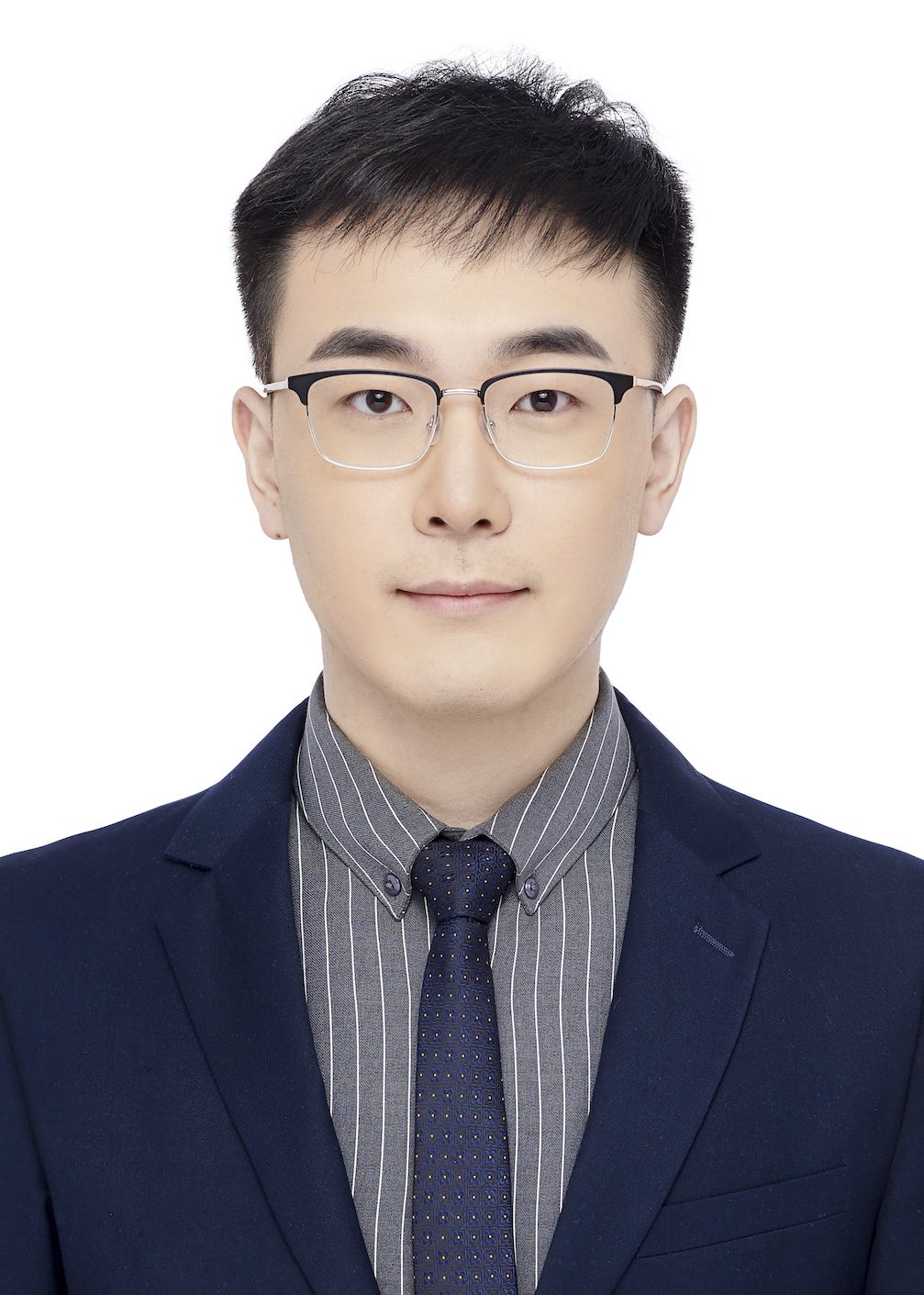}}]{Yongjie
		Duan} received the B.S. degree from the Department of Automation, Tsinghua University, Beijing,
	China, in 2017, where he is currently pursuing the Ph.D. degree under the supervision of Prof.
	Jie Zhou with the Department of Automation. His research interests include fingerprint
	recognition, human-computer interaction, computer vision and pattern recognition.
\end{IEEEbiography}

\begin{IEEEbiography}[{\includegraphics[width=1in,height=1.25in,clip,keepaspectratio]{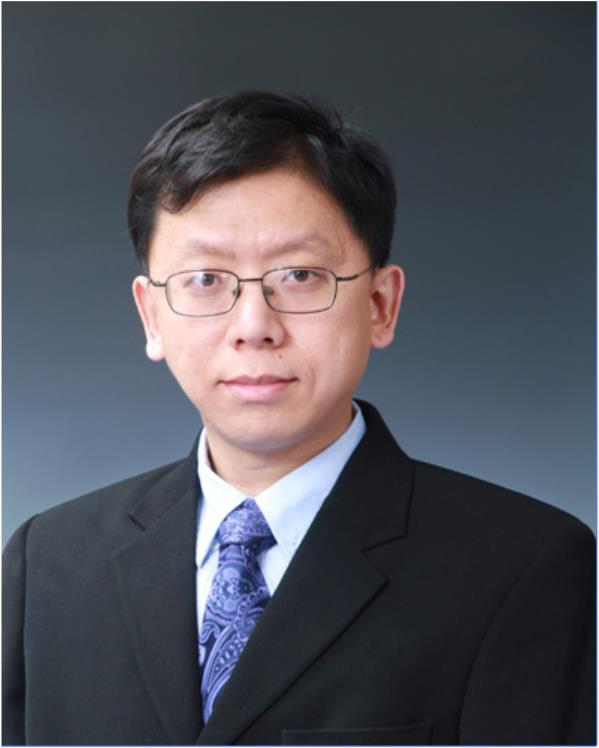}}]{Jianjiang
		Feng} received the B.Eng. and Ph.D. degrees from the School of Telecommunication Engineering,
	Beijing University of Posts and Telecommunications, China, in 2000 and 2007, respectively. From
	2008 to 2009, he was a Post-Doctoral Researcher with the PRIP Laboratory, Michigan State
	University. He is currently an Associate Professor with the Department of Automation, Tsinghua
	University, Beijing. His research interests include fingerprint recognition and computer vision.
\end{IEEEbiography}

\begin{IEEEbiography}[{\includegraphics[width=1in,height=1.25in,clip,keepaspectratio]{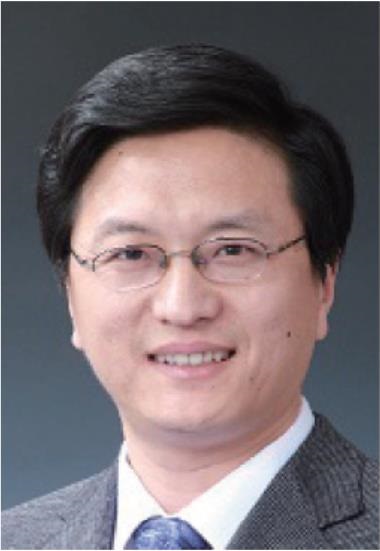}}]{Jie
		Zhou} received the B.S. and M.S. degrees from the Department of Mathematics, Nankai University,
	Tianjin, China, in 1990 and 1992, respectively, and the Ph.D. degree from the Institute of Pattern
	Recognition and Artificial Intelligence, Huazhong University of Science and Technology, Wuhan,
	China, in 1995. From 1995 to 1997, he served as a Post-Doctoral Fellow with the Department of
	Automation, Tsinghua University, Beijing, China. Since 2003, he has been a Full Professor with the
	Department of Automation, Tsinghua University. His research interests include computer vision,
	pattern recognition, and image processing. In recent years, he has authored more than 300 papers in
	peer-reviewed journals and conferences. Among them, more than 100 papers have been published in top
	journals and conferences such as the IEEE Transactions on Pattern Analysis and Machine
	Intelligence, IEEE Transactions on Image Processing, and CVPR. He is an associate editor for the
	IEEE Transactions on Pattern Analysis and Machine Intelligence and two other journals. He received
	the National Outstanding Youth Foundation of China Award. He is a Fellow of the IAPR and a senior
	member of the IEEE.
\end{IEEEbiography}

\end{document}